\documentclass{article}
\usepackage[main,final]{neurips_2026}

\usepackage[utf8]{inputenc}
\usepackage[T1]{fontenc}
\usepackage{hyperref}
\usepackage{url}
\usepackage{booktabs}
\usepackage{amsfonts}
\usepackage{amsmath}
\usepackage{amssymb}
\usepackage{nicefrac}
\usepackage{microtype}
\usepackage{xcolor}
\usepackage{graphicx}
\usepackage{subcaption}   
\usepackage{wrapfig}      

\title{What drives the emergence of symbolic mechanisms in large language models?}

\author{%
  Melody Zixuan Li \\
  Princeton Neuroscience Institute \\
  Princeton University \\
  \texttt{ml8939@princeton.edu} \\
  \And
  Taylor Webb \\
  Princeton Neuroscience Institute \\
  Department of Psychology \\
  Princeton University \\
  \texttt{taylor.w.webb@gmail.com} \\
}

\begin{document}

\maketitle

\begin{abstract}
Recent work has identified a set of emergent symbolic mechanisms that support abstract reasoning in large language models, but it remains unclear what factors drive the emergence of these mechanisms. To address this question, we trained neural networks from scratch on abstract sequence tasks, varying both architectural and data distributional factors, and investigated the effects of these factors on the emergence of symbolic mechanisms. Using representational, attentional, and causal mediation analyses, we first confirmed that transformer-based language models trained from scratch on our task developed symbolic mechanisms. 
We then found that, across multiple axes of data diversity, greater diversity produced stronger symbolic signatures. Surprisingly, the same signatures emerged in modified transformer architectures and even multilayer perceptrons, indicating that architectural inductive biases are not necessary for the development of symbolic mechanisms. We also observed consistent patterns in an entity-attribute binding task. Together, these findings identify data diversity, not architecture, as the primary driver of emergent symbolic computation in neural networks.

\end{abstract}

\section{Introduction}

Can large language models (LLMs) acquire human-like abstract reasoning, and what mechanisms might support this capacity? Many studies have reported that LLMs exhibit strong performance on abstract reasoning tasks~\citep{webb2023emergent}, but debate persists over whether this reflects genuine structured abstraction or surface-level statistical patterns~\citep{mccoy2024embers,lewis2024evaluating}. Symbolic mechanisms have long been hypothesized as necessary for human-like abstraction~\citep{marcus2001algebraic}, yet it has remained unclear whether and how such mechanisms might be implemented and learned by modern neural network models. \citet{yang2025emergentsymbolicmechanismssupport} provided direct evidence for a set of emergent
symbolic mechanisms supporting abstract reasoning in pretrained LLMs, identifying a circuit composed of three 
types of attention-heads: \emph{symbol abstraction heads} that map tokens to abstract role
variables, \emph{symbolic induction heads} that reason over those
variables, and \emph{retrieval heads} that map variables back to tokens.
A key open question concerns the factors that drive the emergence of these mechanisms. Specifically, does the emergence of symbolic computation in LLMs depend on specific inductive biases of the transformer architecture from which these models are constructed, or does it depend on the distribution of the data used to train these models, or some combination of both?

In this work, we sought to address these questions by training neural networks from scratch on a controlled in-context abstract sequence task, similar to the task used to identify symbolic mechanisms in LLMs by \citet{yang2025emergentsymbolicmechanismssupport}. First, we confirmed that transformer language models trained from scratch on this task developed symbol abstraction heads, reproducing some of the key mechanistic signatures identified in pretrained LLMs. We then investigated the extent to which these signatures are driven by data distributional and architectural factors. We found that both mechanistic and behavioral signatures of symbolic computation scaled with \emph{data diversity}, which we varied along four axes: the number of training examples, the vocabulary size, the number of distinct rules and statistically biased training data. On all four axes, greater diversity produced stronger symbolic signatures. We also found an interaction between data diversity and training dynamics: Models trained with small vocabularies and few rules initially solved the task using a non-symbolic solution based on first-order statistics, but eventually developed symbolic mechanisms with further training, following a pattern similar to grokking \citep{power2022grokking,nanda2023progress}. By contrast, we found that the emergence of symbolic mechanisms did not depend on architectural inductive biases. The same mechanistic and behavioral signatures emerged in a set of transformer architectures modified to remove specific inductive biases relevant to symbolic computation and even in a multilayer perceptron (MLP) architecture with much weaker inductive biases. Finally, to test whether these findings extend beyond the abstract sequence task, we introduced a second task that required in-context binding of entities to attributes, where we found that some of the same data-diversity factors governed the emergence of the binding-ID mechanism identified by \citet{feng2024binding}, and that a signature of this mechanism emerged in MLPs.


We make the following contributions:

\begin{enumerate}
    \item  We found that transformers trained from scratch on abstract sequence tasks developed a two-stage symbolic mechanism. Symbol abstraction heads inferred the rule, and a rotation mechanism applied it to the query (Section~\ref{sec:solution}).
    \item Varying diversity along four axes, we found that it governed whether this mechanism emerged, as shown by causal interventions and out-of-distribution generalization (Section~\ref{sec:diversity}).
    \item  Surprisingly, our experiments revealed that modified transformers, and even plain MLPs, exhibited the same signatures of symbolic computation (Section~\ref{sec:architecture}).
    \item  We further observed that on entity-attribute binding, data diversity likewise determined whether a binding-ID mechanism emerged, while architecture had no impact (Section~\ref{sec:binding}).
\end{enumerate}
\section{Analysis of the Solution Learned by the Model}
\label{sec:solution}

\subsection{Architecture of the Learned Solution}

\begin{figure}[t]
    \centering
    \includegraphics[width=0.66\linewidth]{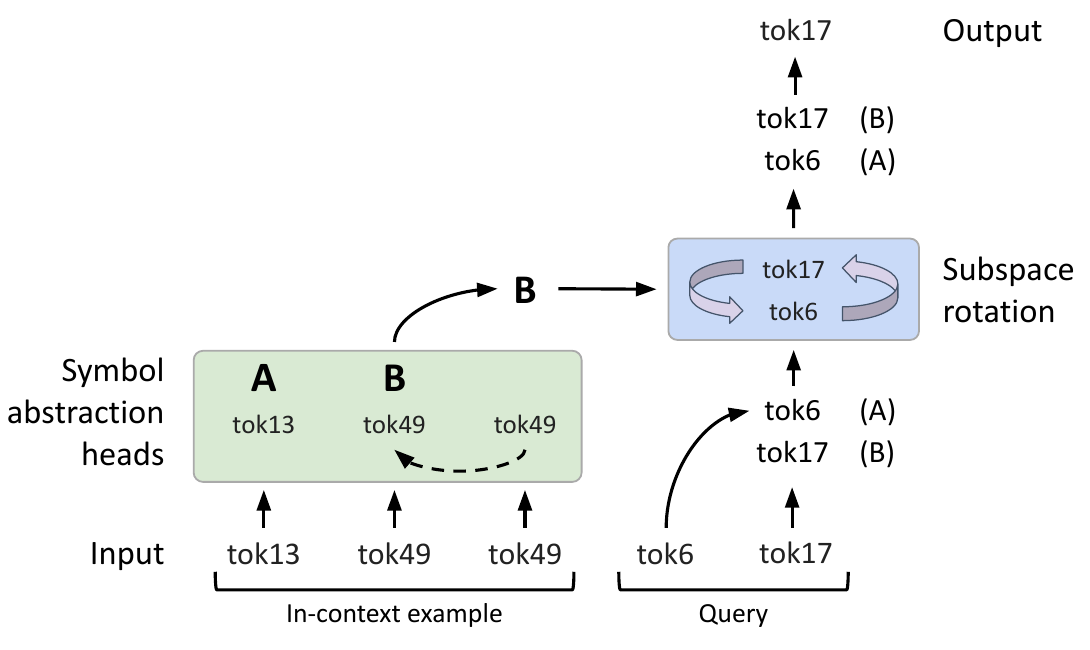}
    \caption{\textbf{Illustration of Symbolic Mechanisms Learned by Model.} Problems consist of an
in-context example governed by either an ABA or ABB (pictured) rule constructed with arbitrary tokens, followed by an incomplete query example that must be completed according to the same rule.
Symbol abstraction heads process the in-context example, identifying the symbol (A or B) associated with the third token. Query tokens are copied into role-specific (A vs. B) subspaces of the residual stream at the final position, and a rule-dependent (ABA vs. ABB) rotation is performed to gate the appropriate token (corresponding to A vs. B) into the output subspace.}
    \label{fig:architecture}
\end{figure}

We trained transformer language models to perform an abstract sequence task similar to the one used by \citet{yang2025emergentsymbolicmechanismssupport} to identify emergent symbolic mechanisms in pretrained LLMs. In our task, an in-context example governed by either an ABA or ABB rule, constructed from arbitrary tokens, was followed by a query consisting of two tokens. We trained models to predict the token that would complete the query example according to the rule inferred from the in-context example.
We first identified the emergent circuit these models learned to solve the task (Figure~\ref{fig:architecture}). The circuit consisted of two major components: 1) \emph{symbol abstraction heads} that computed invariant symbolic representations of the rule governing the in-context example, and 2) a rule-dependent \emph{rotation} operation in the residual stream at the final position that gated the appropriate token (corresponding to A vs. B, depending on the rule) into the output subspace for the model. The following sections describe the evidence for these mechanisms.


\begin{figure}[t]
  \centering


  \hfill
  \begin{subfigure}[t]{0.33\linewidth}
    \centering
    \includegraphics[width=\linewidth]{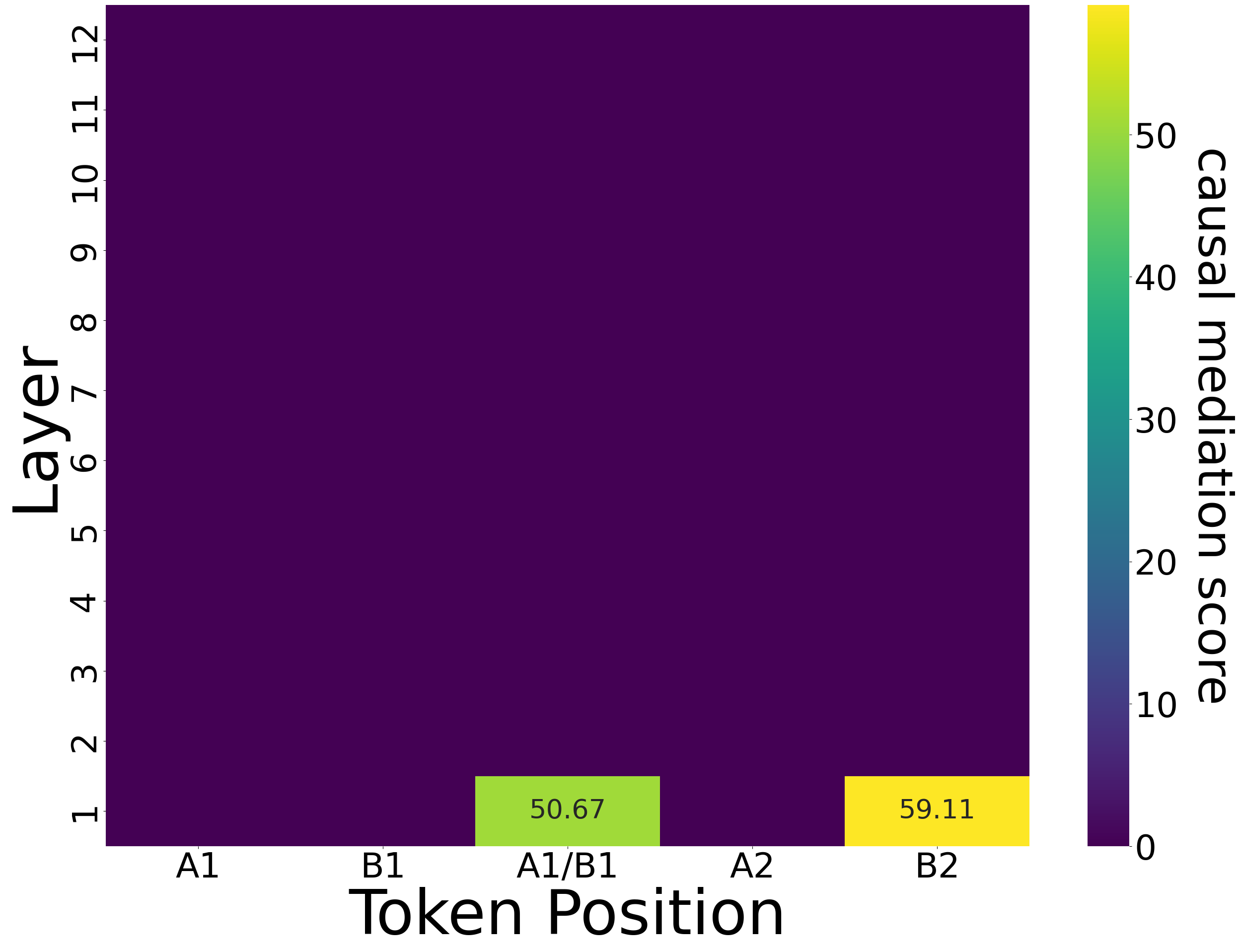}
    \caption{Layer-wise CMA}
    \label{fig:cma_layer}
  \end{subfigure}
  \hfill
  \begin{subfigure}[t]{0.33\linewidth}
    \centering
    \includegraphics[width=\linewidth]{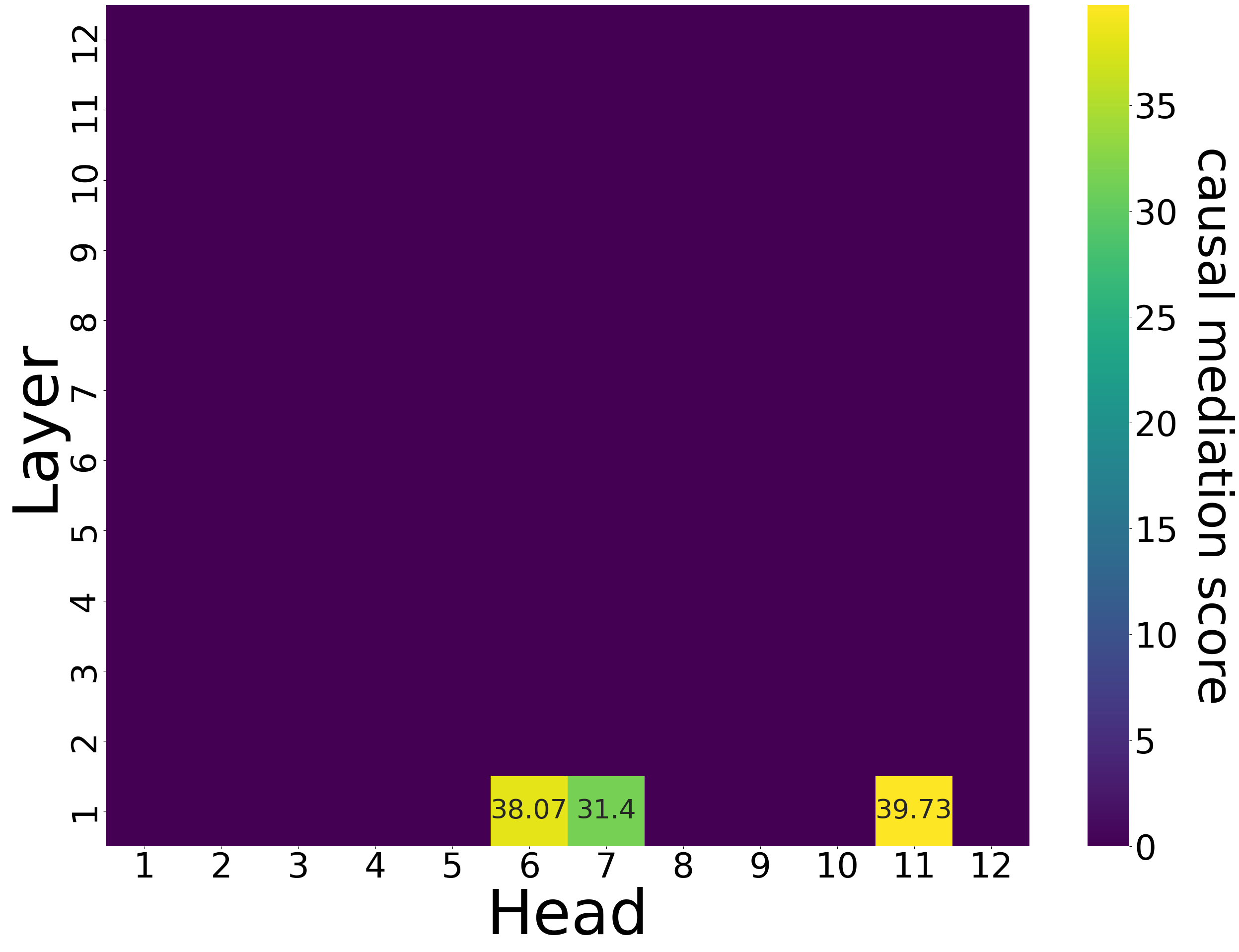}
    \caption{Head-wise CMA}
    \label{fig:cma_head}
  \end{subfigure}
   \begin{subfigure}[t]{0.32\linewidth}
    \centering
    \includegraphics[width=\linewidth]{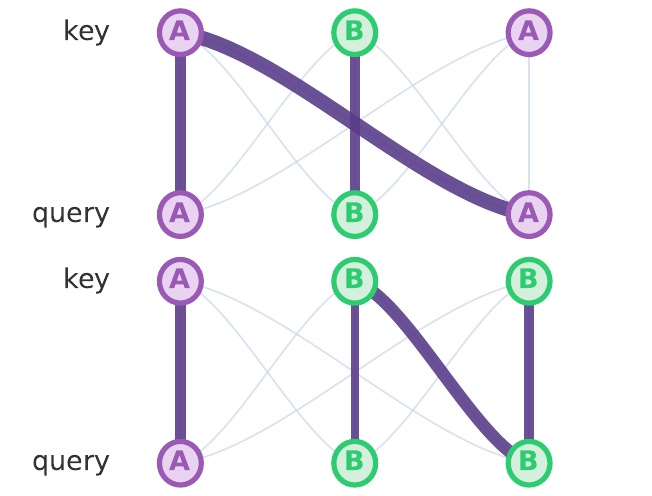}
    \caption{Attention patterns}
    \label{fig:cma_attn}
  \end{subfigure}

  \vspace{0.8em}

  \begin{subfigure}[t]{0.32\linewidth}
    \centering
    \includegraphics[width=\linewidth]{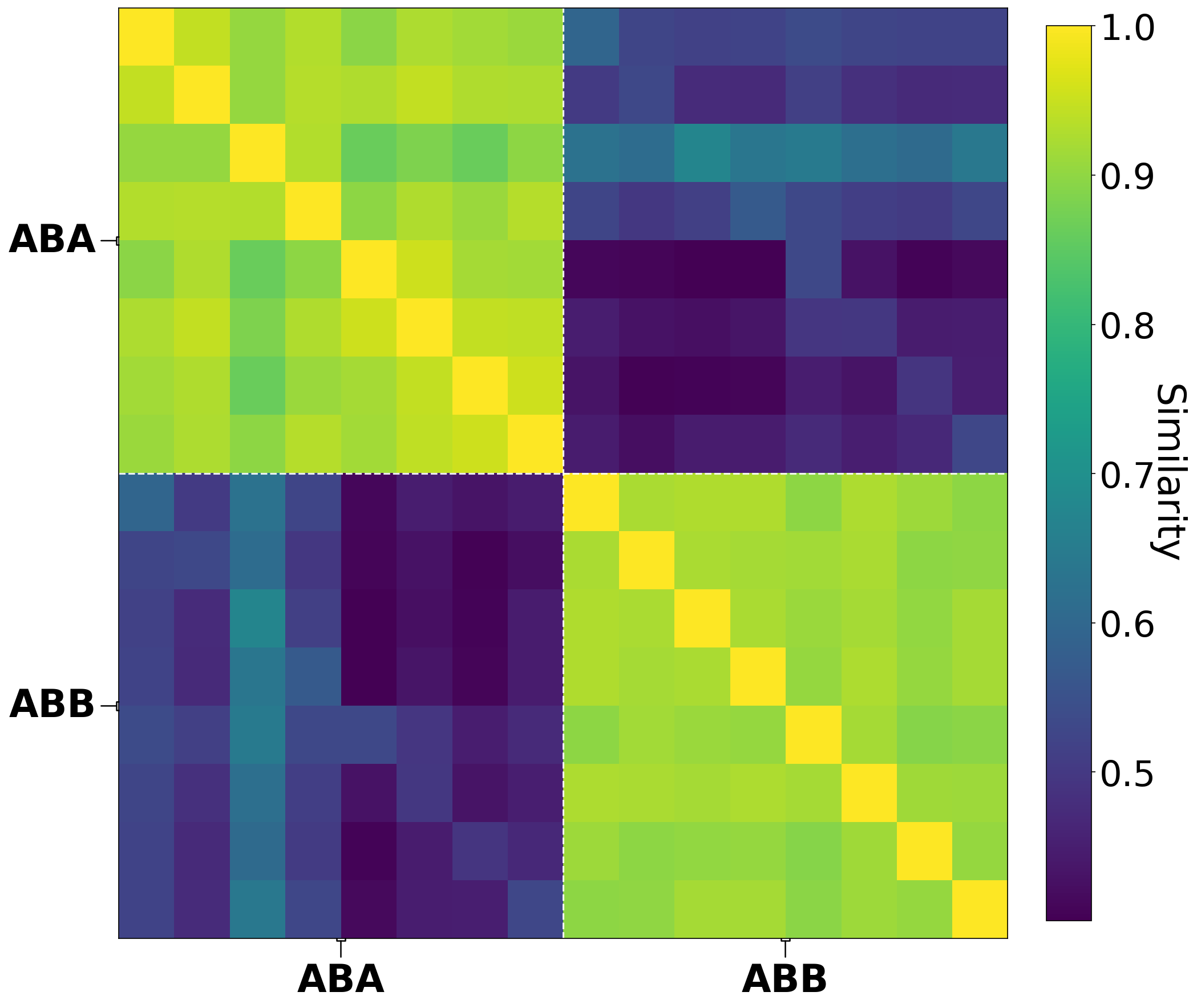}
    \caption{CMA-weighted average RSA}
    \label{fig:rsa}
  \end{subfigure}
  \hfill
  \begin{subfigure}[t]{0.33\linewidth}
    \centering
    \includegraphics[width=\linewidth]{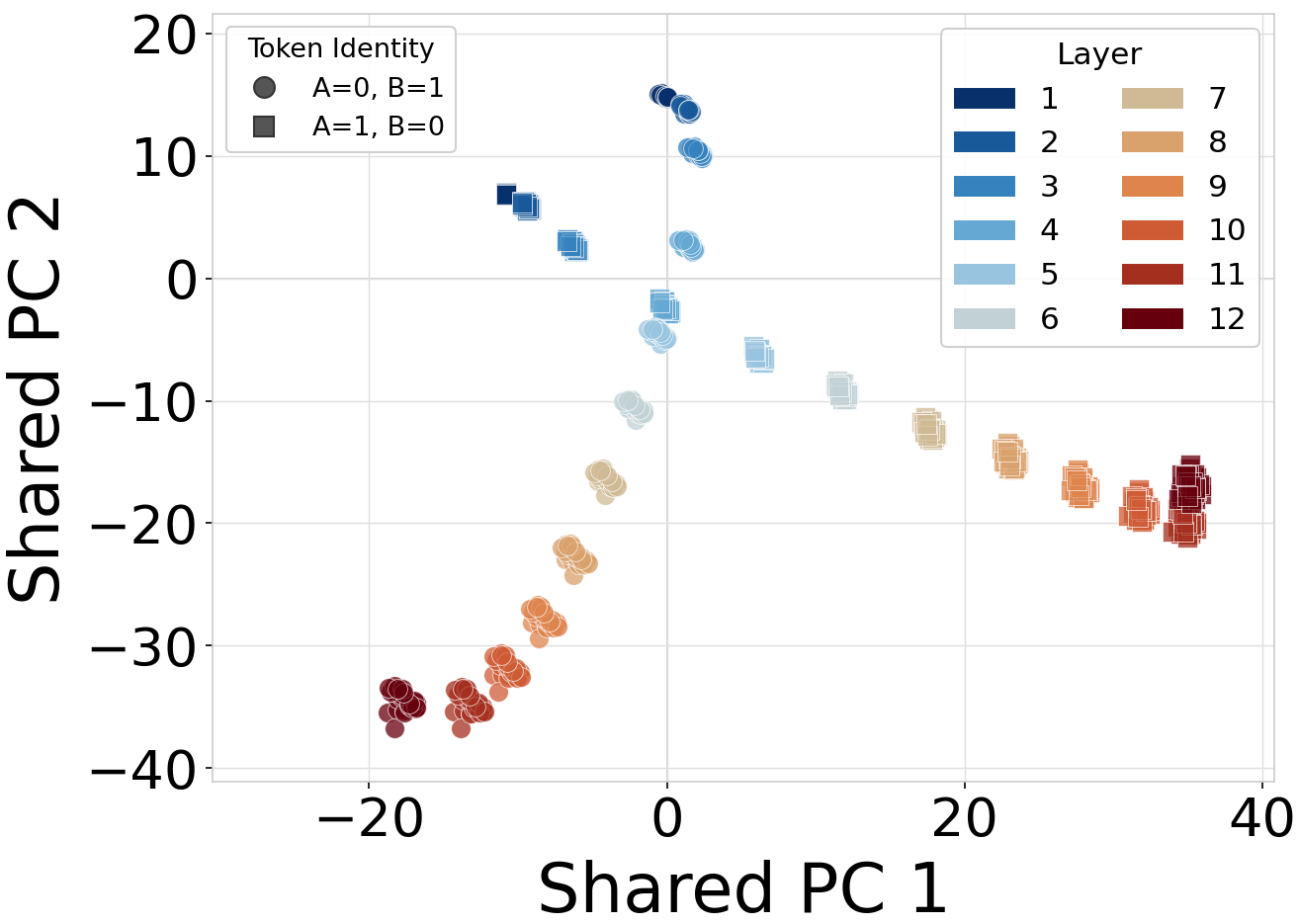}
    \caption{PCA: ABA}
    \label{fig:pca_early}
  \end{subfigure}
  \hfill
  \begin{subfigure}[t]{0.33\linewidth}
    \centering
    \includegraphics[width=\linewidth]{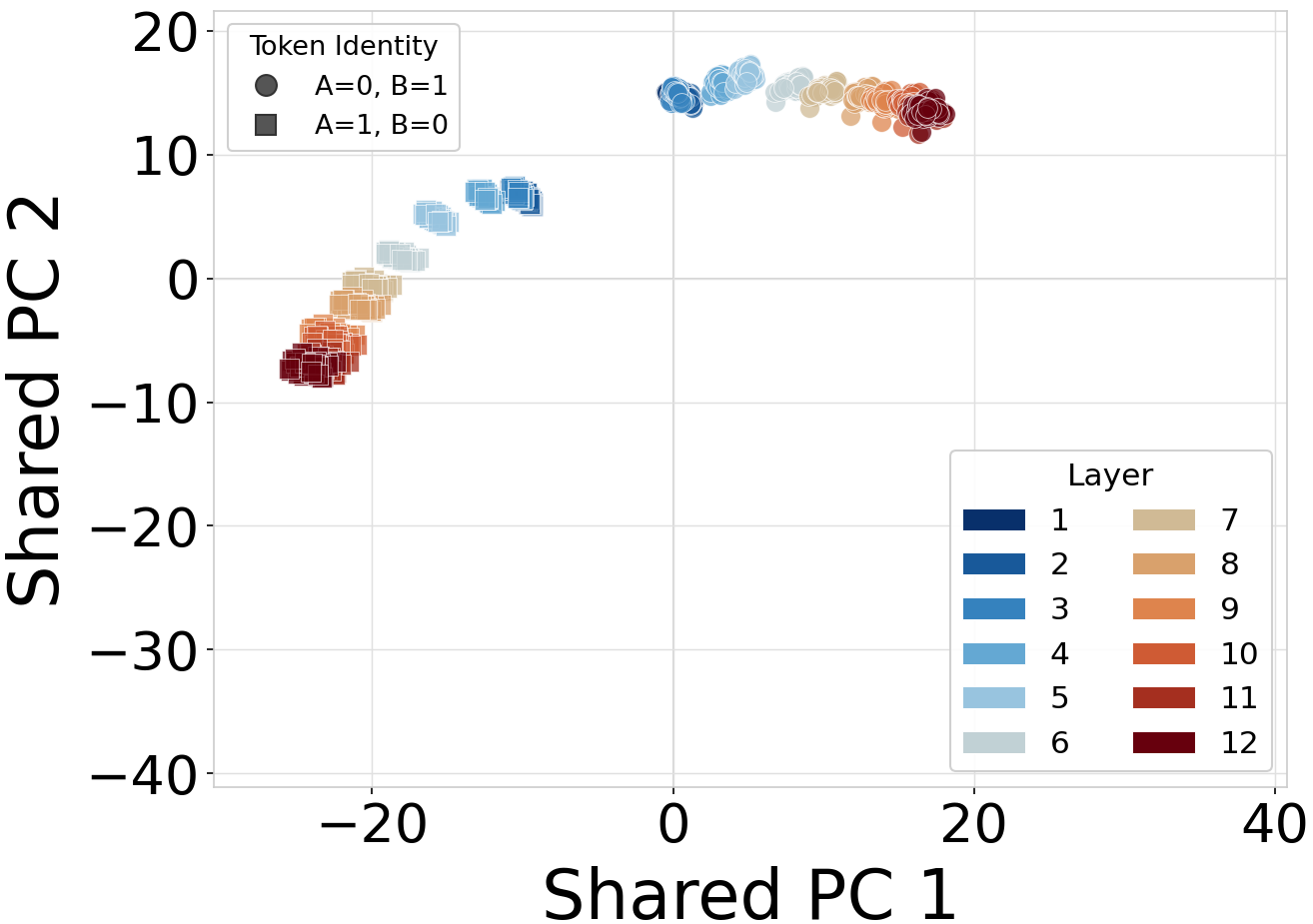}
    \caption{PCA: ABB}
    \label{fig:pca_final}
  \end{subfigure}

\caption{\textbf{Evidence for Symbol Abstraction Heads and Rotation Mechanism Learned by Model.}
    \textbf{(a)}~Layer-wise causal mediation analysis localizes
    symbolic encoding to layer~1.
    \textbf{(b)}~Head-wise causal mediation analysis identifies symbol abstraction heads.
    \textbf{(c)}~Query--key attention of the symbol abstraction heads.
    \textbf{(d)}~RSA confirms that identified heads encode abstract role
    identity rather than token surface form.
    \textbf{(e)}--\textbf{(f)}~Residual-stream PCA reveals
    a layer-wise rotation from abstract role representation to output
    prediction.}
  \label{fig:cma}
\end{figure}
\subsection{Evidence for Symbol Abstraction Heads}
\label{sec:sah}

Symbol abstraction heads are attention heads that convert input tokens to invariant symbol-like representations, based on their relations to other tokens \citep{yang2025emergentsymbolicmechanismssupport}. They are an emergent instance of the Abstractor architecture proposed by \citet{altabaa2024abstractors}. In  the specific case of our task, these heads convert input tokens to invariant representations of abstract symbols (i.e., A or B) based on the relationship to the other tokens within an in-context example (e.g., if the third token is the same as the second token, it will be assigned the symbol B). We identified a set of heads in the first layer of the model that reproduced the key signatures (attentional, representational, and causal) of symbol abstraction heads found in LLMs.

\paragraph{Causal Mediation Analysis.}
To determine whether our model developed symbol abstraction heads, we performed causal mediation analysis (CMA) ~\citep{pearl2022direct,wang2022interpretability,todd2024function}, patching internal activations between contexts with different rules (ABA vs. ABB) and measuring the effect on the model's outputs. Importantly, context pairs were created in which the rule differed, but the final token in the in-context example was the same (e.g., `tok13 tok49 tok49', instantiating an ABB rule vs. `tok49 tok13 tok49', instantiating an ABA rule, but both ending with `tok49'), ensuring that any effect observed for patching activations at this token position would reflect encoding of an abstract symbol rather than merely encoding input tokens. When patching activations for the output of all attention heads at a particular layer, we found an effect concentrated in layer 1 for the expected position, which is the third item (A1/B1) in the in-context example (Figure~\ref{fig:cma_layer})\footnote{Note that we also found an effect at the final position in the sequence (B2), but this does not unambiguously reflect the abstract rule since the input token also differed at this position.}. When patching activations for the outputs of individual attention heads at this token position, we found a set of three attention heads that produced the expected effect (Figure~\ref{fig:cma_head}), thus identifying these heads as candidate symbol abstraction heads.

\paragraph{Attention Analysis.}

To further confirm the presence of symbol abstraction heads, we analyzed the attention patterns for the identified heads. Consistent with the predicted behavior of symbol abstraction heads, we found that attention was directed from the final token in each in-context example to the previous token that shared the same symbol (i.e., from the third token to the first token for ABA rules, and from the third token to the second token for ABB rules) (Figure~\ref{fig:cma_attn}).


\paragraph{Representational Similarity Analysis.}
To test whether the identified heads encode truly invariant symbolic representations rather than token identity, we performed representational similarity analysis (RSA)~\citep{kriegeskorte2008representational}. For each identified head, we computed the pairwise similarity between the outputs of that head across pairs of contexts, constructing a similarity matrix reflecting the pattern of similarity across many contexts. Figure~\ref{fig:rsa} shows a weighted average of this matrix, weighted by the CMA scores used to identify these heads. The results confirmed that the outputs of these heads had a high degree of similarity for pairs of contexts that shared the same rule (upper left and lower right quadrants) regardless of the specific tokens used to instantiate those rules, confirming that these heads encoded abstract rules invariantly. The identified attention heads therefore reproduced the key signatures of symbol abstraction heads found in pretrained LLMs.


\subsection{Evidence for Rule-Dependent Subspace Rotation}
\label{sec:rotation}

After the rule is inferred from the in-context example, it must be applied to the query to generate the final token. Rather than employing symbolic induction heads and retrieval heads as in pretrained LLMs \citep{yang2025emergentsymbolicmechanismssupport}, we found that this was accomplished in our model via a rule-dependent subspace rotation in the residual stream at the final position.  Specifically,  the A and B tokens for the query example were stored in role-specific subspaces in early layers, and these subspaces were then rotated in a rule-dependent manner (e.g., the subspaces are rotated only for ABA rules, not for ABB rules) such that the appropriate completion token (A or B, depending on the rule) is available to be read out from a particular subspace in the later layers of the model.

\paragraph{Shared PCA.}
We confirmed the presence of this rotation-based strategy by performing PCA across all layers jointly on residual-stream activations at the final position, thus constructing a shared subspace in which layer-wise trajectories can be compared directly.
Problems were constructed so that the query token was drawn from a fixed pair of tokens, allowing token identity and abstract role (A vs. B) to be dissociated across trials. 
Figure~\ref{fig:pca_early} and Figure~\ref{fig:pca_final} show resulting trajectories for ABA and ABB rules, with color encoding layer
depth (blue = early, red = late) and marker shape encoding token
identity.
In early layers, representations clustered by token identity regardless of rule. Across successive layers, the trajectories followed an arc-shaped path in the shared PCA space, converging by the final layer to a configuration that is separable by the correct output token rather than by abstract role. These results are consistent with a near-linear rotation that progressively remaps the abstract role representation into the appropriate output token embedding, depending on the rule.

\paragraph{Decoding role information.}
We trained linear decoders on the residual stream at the final sequence position and evaluated whether they generalized across rules (ABA to ABB and vice versa). In early layers, decoders transferred based on symbol identity, independent of the rule, whereas in later layers, they transferred based on the rule-dependent correct answer, which differs between the two rules. This transition occurred in the middle layers, consistent with the PCA trajectories. Further details are provided in Appendix~\ref{app:decoder}.
\section{Data Diversity Drives the Emergence of Symbolic Mechanisms}
\label{sec:diversity}

\subsection{Experimental Setup}

We varied data diversity along four axes to evaluate its impact on the emergence of symbol abstraction heads. All models used the same hyperparameters as GPT-2 Small \citep{radford2019language}. The first data diversity axis was the \textbf{number of training examples} (from 200 to 200k). The second was \textbf{role-token bias} ($0\%$ to $100\%$), defined as the fraction of vocabulary tokens consistently associated with a single role (A or B) during training. These associations were then reversed at test time. The third was \textbf{vocabulary size} ($V$ from 50 to 10k). The fourth was the \textbf{number of distinct rules} (2 or 10). Across all diversity axes, we hypothesized that increased diversity would reduce the viability of token-level or statistical shortcuts and would therefore encourage models to learn a solution based on genuine relational structure. To compare the emergence of symbolic mechanisms across these settings, we used the concept intervention described below and measured its efficacy. 
For role-token bias, we additionally evaluated out-of-distribution (OOD) generalization behaviorally. The specific parameter values, data construction procedures, and training details are provided in Appendix~\ref{app:details}.
\subsection{Concept Interventions}
\label{sec:ci}
\begin{figure}[t]
  \centering

  \begin{minipage}[t]{0.38\linewidth}
    \vspace{0pt}
    \subcaptionbox{Concept intervention pipeline\label{fig:ci_pipeline}}{
      \includegraphics[width=\linewidth]{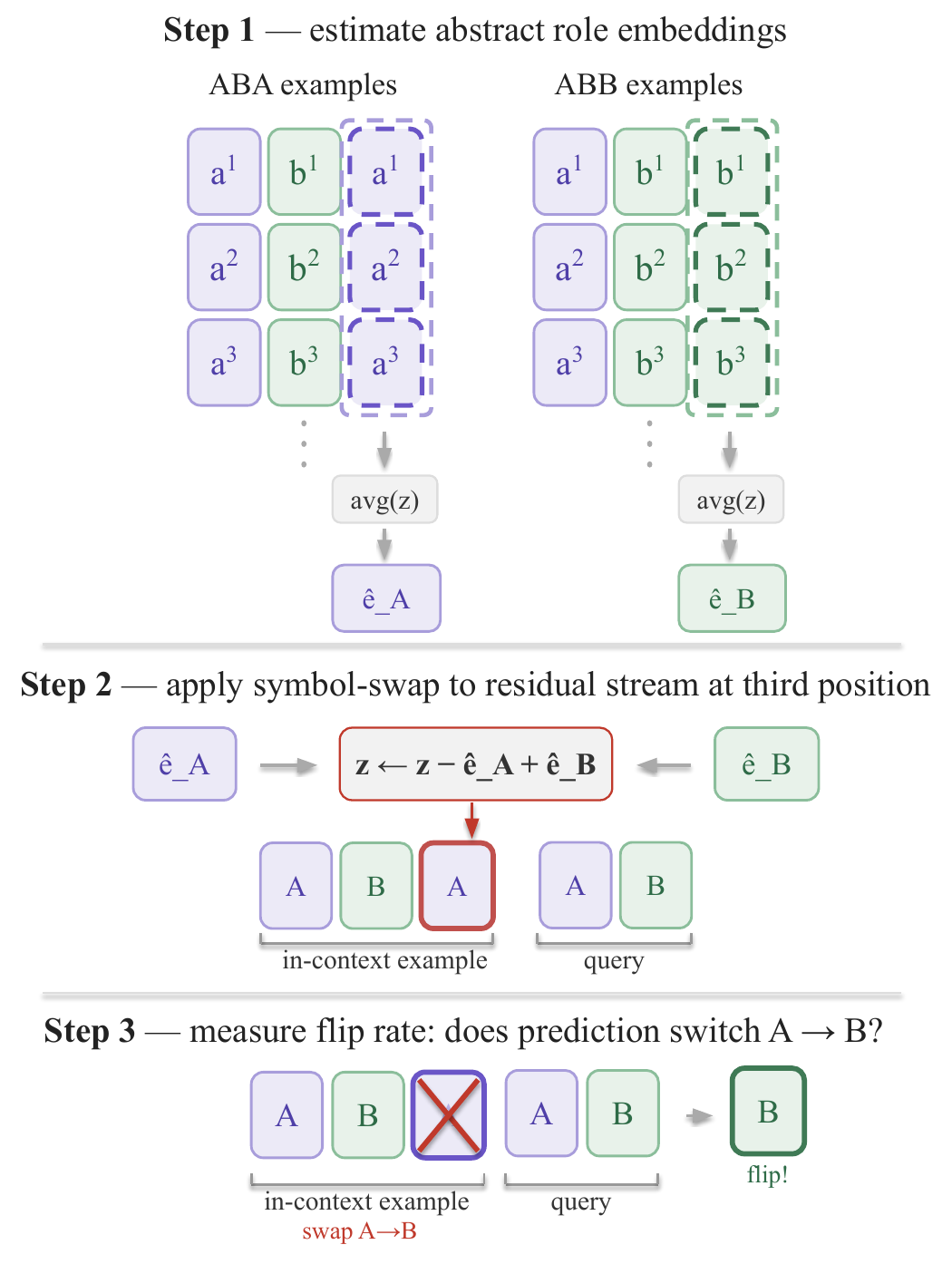}
    }
  \end{minipage}
  \hfill
  \begin{minipage}[t]{0.3\linewidth}
    \vspace{0pt}
    \subcaptionbox{ \# training examples\label{fig:fliprate_nsamples}}{
      \includegraphics[width=\linewidth]{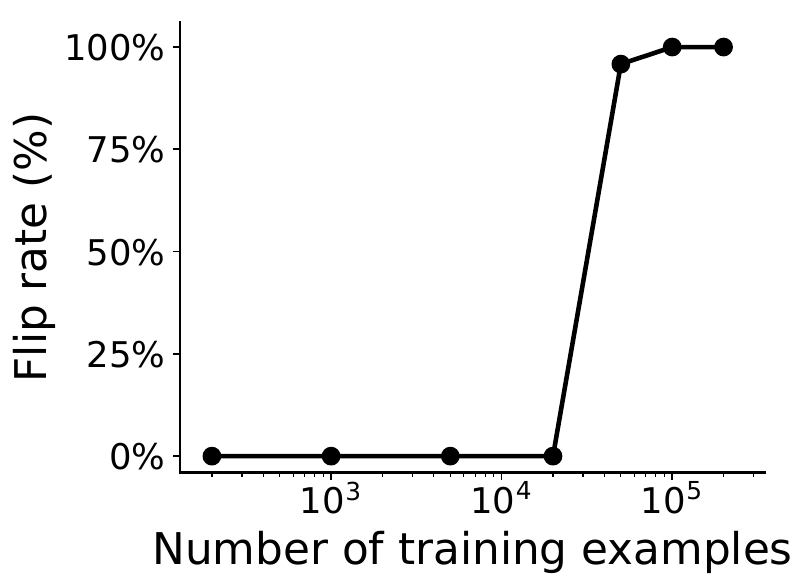}
    }

    \vspace{0.6em}

    \subcaptionbox{ \ role-token bias\label{fig:fliprate_bias}}{
      \includegraphics[width=\linewidth]{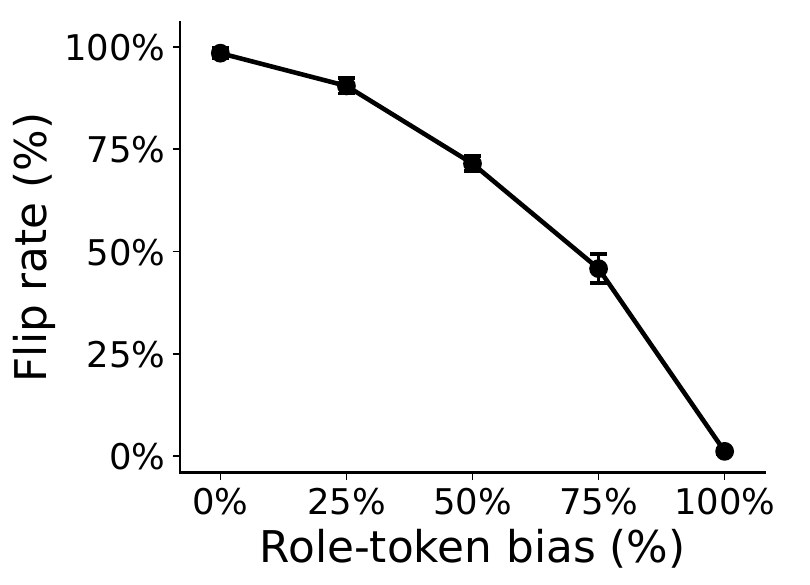}
    }
  \end{minipage}
  \hfill
  \begin{minipage}[t]{0.3\linewidth}
    \vspace{0pt}
    \subcaptionbox{vocab.\ size\label{fig:fliprate_vocab}}{
      \includegraphics[width=\linewidth]{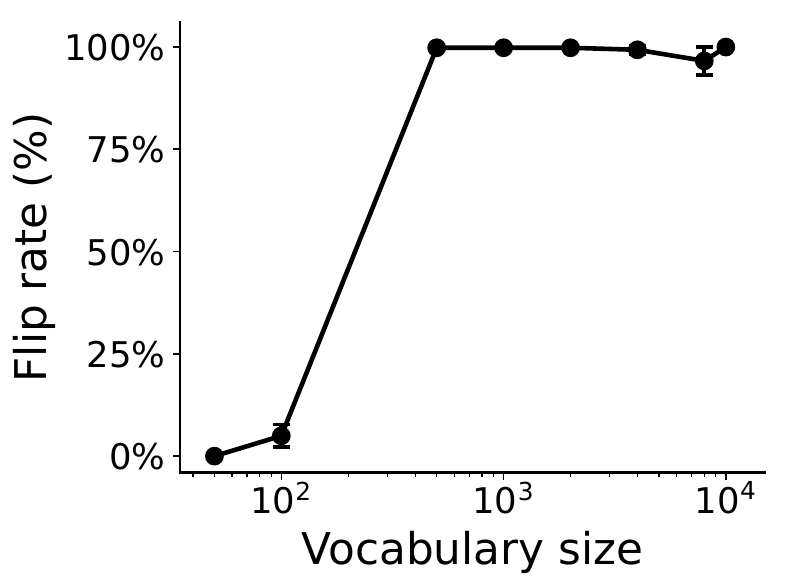}
    }
        \vspace{0.6em}
        
    \subcaptionbox{\# rules \label{fig:fliprate_rules}}{
      \includegraphics[width=\linewidth]{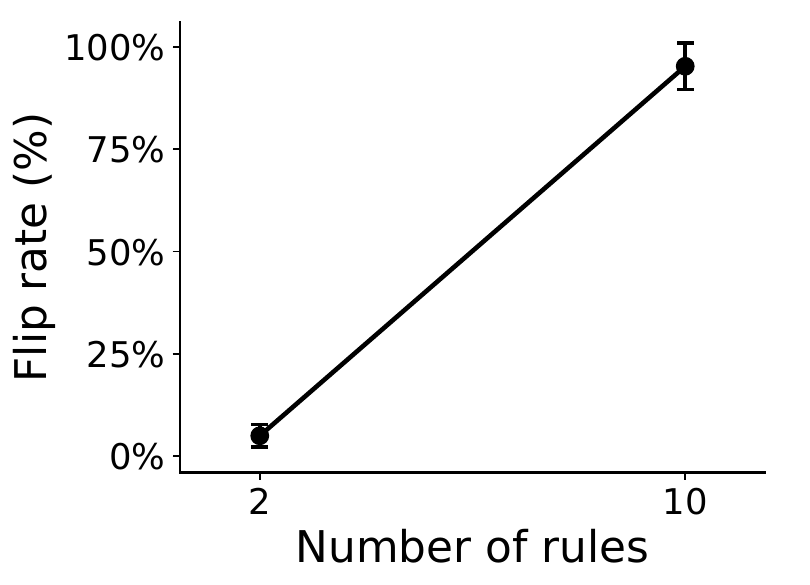}
    }

  \end{minipage}

\caption{\textbf{Assessing the Impact of Data Diversity on Emergent Symbol Processing.}
    \textbf{(a)}~Concept intervention pipeline. Role embeddings $\hat{\mathbf{e}}_A$ and $\hat{\mathbf{e}}_B$ are estimated from residual activations and used to swap the role at a target position. The flip rate is the fraction of prompts whose prediction switches to the answer of the opposite rule. \textbf{(b)}~Flip rate increases with the number of unique training examples,~\textbf{(c)} decreases as role-token bias increases, and~\textbf{(d)} increases with vocabulary size and~\textbf{(e)} the number of rules. Error bars show the standard deviation across 3 seeds.}
  \label{fig:concept_intervention}
\end{figure}

Let $\mathbf{z}_{t,\ell}$ denote the residual stream at position $t$ and layer $\ell$.
Let $t^\star$ be the third token of the in-context example, which is the A symbol in an ABA example and the B symbol in an ABB example.
For each layer we define the role embeddings as mean activations at $t^\star$,
\begin{equation}
  \hat{\mathbf{e}}_{A,\ell} = \mathbb{E}_{\mathrm{ABA}}\!\left[\mathbf{z}_{t^\star,\ell}\right],
  \qquad
  \hat{\mathbf{e}}_{B,\ell} = \mathbb{E}_{\mathrm{ABB}}\!\left[\mathbf{z}_{t^\star,\ell}\right],
\end{equation}
estimated from $n$ matched prompts of each type. To swap the role on a test prompt, we change the activation at $t^\star$ only. For an ABA prompt,
\begin{equation}
  \mathbf{z}_{t^\star,\ell} \;\leftarrow\; \mathbf{z}_{t^\star,\ell} - \hat{\mathbf{e}}_{A,\ell} + \hat{\mathbf{e}}_{B,\ell}
  \quad \text{for every layer } \ell \text{ simultaneously},
\end{equation}
and for an ABB prompt we exchange $\hat{\mathbf{e}}_{A,\ell}$ and $\hat{\mathbf{e}}_{B,\ell}$. 
When applied to an ABA sequence, the intervention should shift the model's predictions toward those consistent with ABB. We applied the intervention at the third position of the in-context example, the token that determines whether the example follows ABA or ABB. We measured the \emph{flip rate}, the fraction of trials in which the intervened model outputs the answer that the opposite rule would give. We used the flip rate of this intervention as our primary measure of the extent to which the model possesses invariant, causally relevant symbolic representations. (Figure~\ref{fig:ci_pipeline}). 


We began by varying the number of unique training examples. Flip rate was near $0\%$ with $200$ examples and reached $100\%$ from $100$k examples (Figure~\ref{fig:fliprate_nsamples}). In these low-diversity conditions the models did not generalize: training accuracy converged, but test accuracy did not (Appendix Figure~\ref{fig:tc_ntrain}). To test whether symbolic mechanisms depend on diversity even when models do generalize, we turned to role-token bias. During training, a fraction $b$ of the vocabulary was role-exclusive: half of these tokens appeared only as A and half only as B, while the remaining $(1{-}b)V$ tokens can fill either role. At $b{=}100\%$ a token's identity fully determines its role, so the training distribution can be solved by learning token--role associations alone. At test time, we reversed the roles of the exclusive tokens. All models generalized in-distribution at every bias level, including $100\%$ bias, with in-distribution accuracy near ceiling (Appendix Figure~\ref{fig:tc_bias}). Flip rate nevertheless fell from $98\%$ at $0\%$ bias to near $0\%$ at $100\%$ bias (Figure~\ref{fig:fliprate_bias}). For vocabulary size, flip rate was $0\%$ at $V{=}50$, $5\%$ at $V{=}100$, and close to $100\%$ from $V{=}500$ upward (Figure~\ref{fig:fliprate_vocab}). At $V{=}100$, models generalized but had a low flip rate (consistent with a lack of symbolic mechanisms), so we asked whether training models on more rules would induce a higher flip rate at this vocabulary size. With $V{=}100$, flip rate was $5\%$ with $2$ rules and $95\%$ with $10$ rules (Figure~\ref{fig:fliprate_rules}). 

Finally, we found an interaction between data diversity and training dynamics. For all experiments described above, we measured flip rate at the point in training when models first reached an accuracy threshold of $95\%$. For models trained on a small vocabulary size ($V{=}50$ and $V{=}100$), we found that training models well past the point of convergence eventually led to a sharp increase in the flip rate, consistent with an initial period where models employed a non-symbolic but generalizable solution followed by a later period in which these models eventually developed symbolic mechanisms (Appendix~\ref{app:diversity_dynamics}). Together, these results show that greater data diversity promotes the efficient development of symbolic mechanisms.

\subsection{Out-of-Distribution Generalization}
\label{sec:ood}
\begin{wrapfigure}{r}{0.44\linewidth}
  \centering
  \vspace{-1em}
  \includegraphics[width=\linewidth]{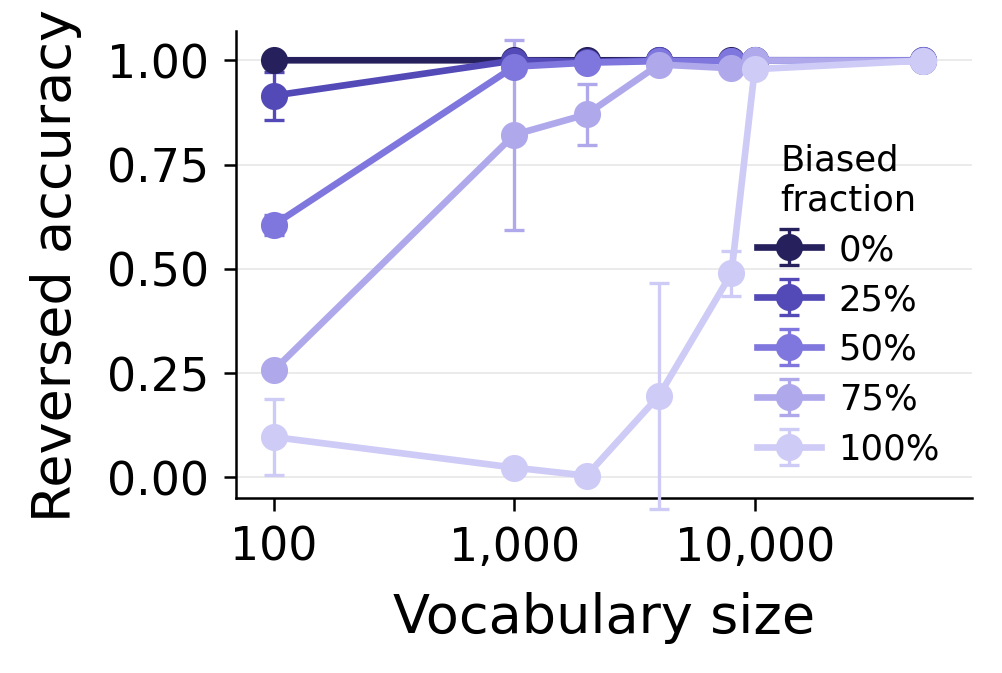}
  \caption{Out-of-distribution generalization}
  \label{fig:bias_acc}
  \vspace{-1em}
\end{wrapfigure}

For models trained on different levels of role-token bias, we also measured accuracy on test prompts in which the token–role assignments from training were reversed (Figure~\ref{fig:bias_acc}). A model that relies on first-order token statistics should fail on these tests, because the associations it learned now lead to the wrong answer. By contrast, a model that has learned the rule should be unaffected. Consistent with this, we found that models trained on a small vocabulary size ($V{=}100$, $2$
rules), generalized well when trained with $0\%$ role-token bias (i.e., they generalized well in-distribution), but were unable to generalize when trained with $100\%$ role-token bias (i.e., they could not generalize out-of-distribution). By contrast, models trained with a larger vocabulary size generalized well at all bias levels (i.e., they generalized well both in-distribution and out-of-distribution). These results suggest that, in addition to the mechanistic signature tracked by our concept interventions, data diversity also drove out-of-distribution generalization at the behavioral level, thus conferring a key benefit of symbolic computation.


\section{Symbolic Mechanisms Do Not Depend on Architectural Inductive Biases}
\label{sec:architecture}
\begin{figure}[t]
    \centering
    \begin{subfigure}[t]{0.49\textwidth}
        \centering
        \includegraphics[width=\linewidth]{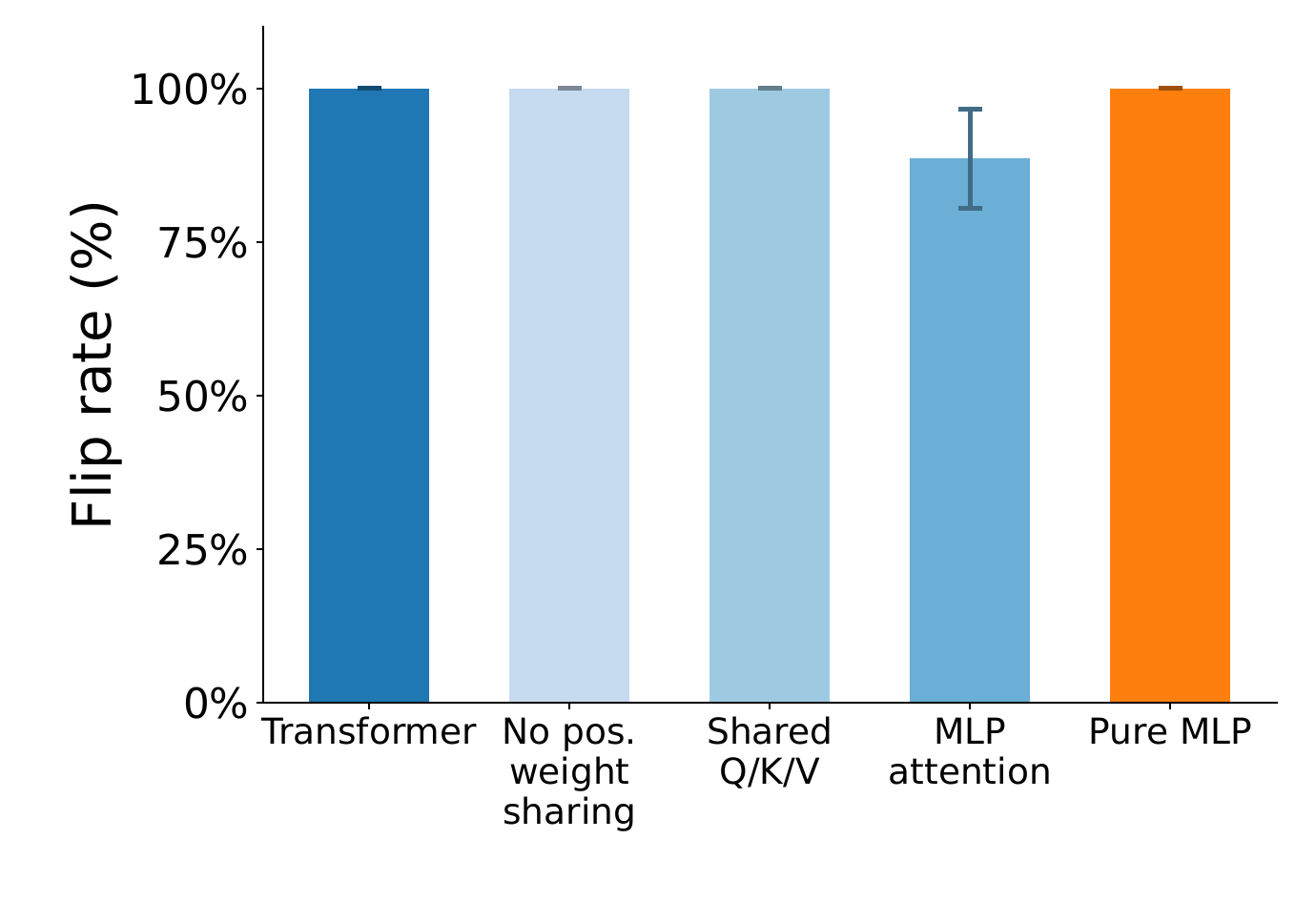}
        \caption{Architectural variants}
        \label{fig:A}
    \end{subfigure}
     \hfill
    \begin{subfigure}[t]{0.49\textwidth}
        \centering
        \includegraphics[width=\linewidth]{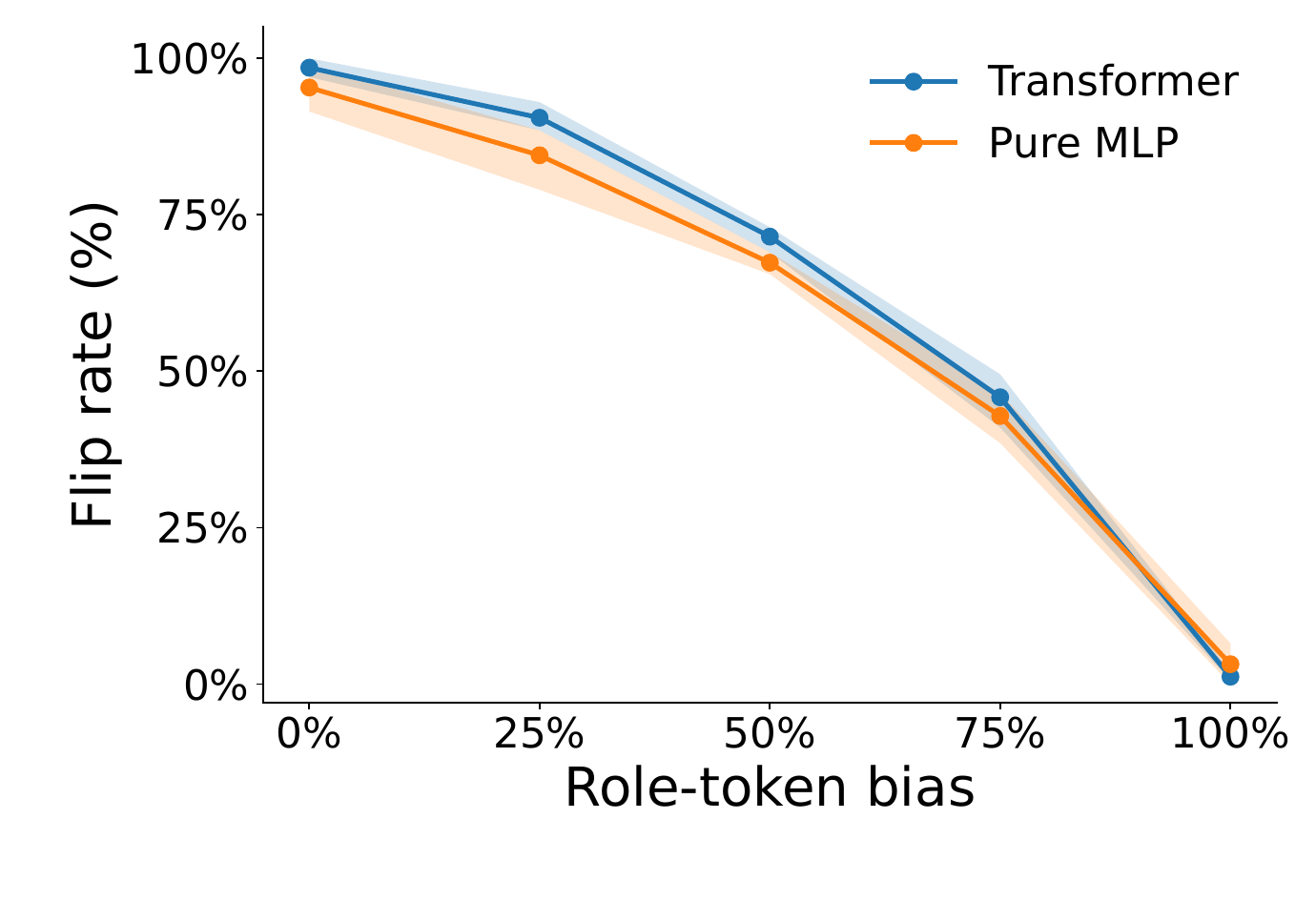}
        \caption{MLP flip rate}
        \label{fig:B}
    \end{subfigure}
    \hfill
    \begin{subfigure}[t]{0.49\textwidth}
        \centering
        \includegraphics[width=\linewidth]{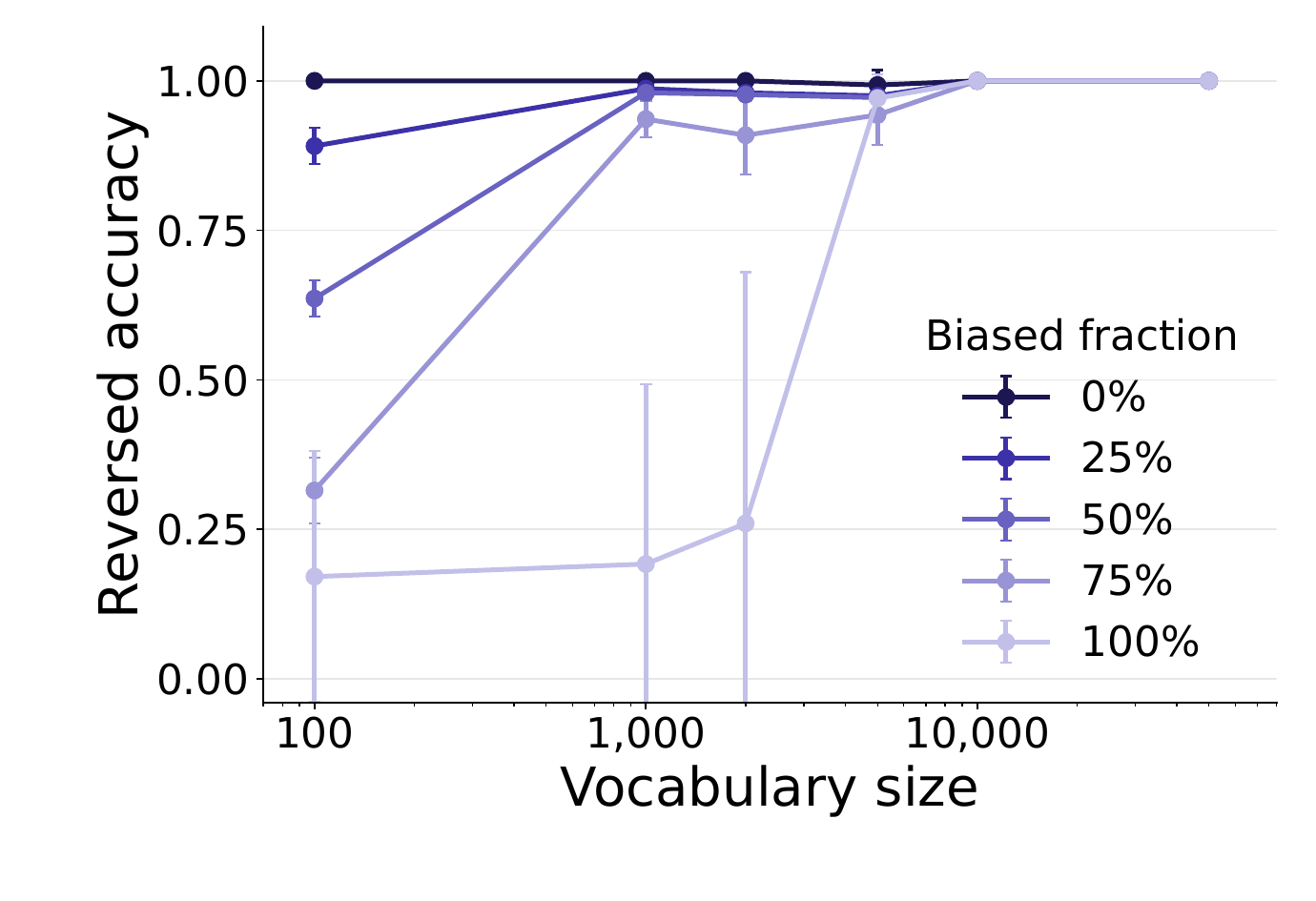}
        \caption{MLP out-of-distribution generalization}
        \label{fig:C}
    \end{subfigure}
    \vspace{0.5em}
    \begin{subfigure}[t]{0.49\textwidth}
        \centering
        \includegraphics[width=\linewidth]{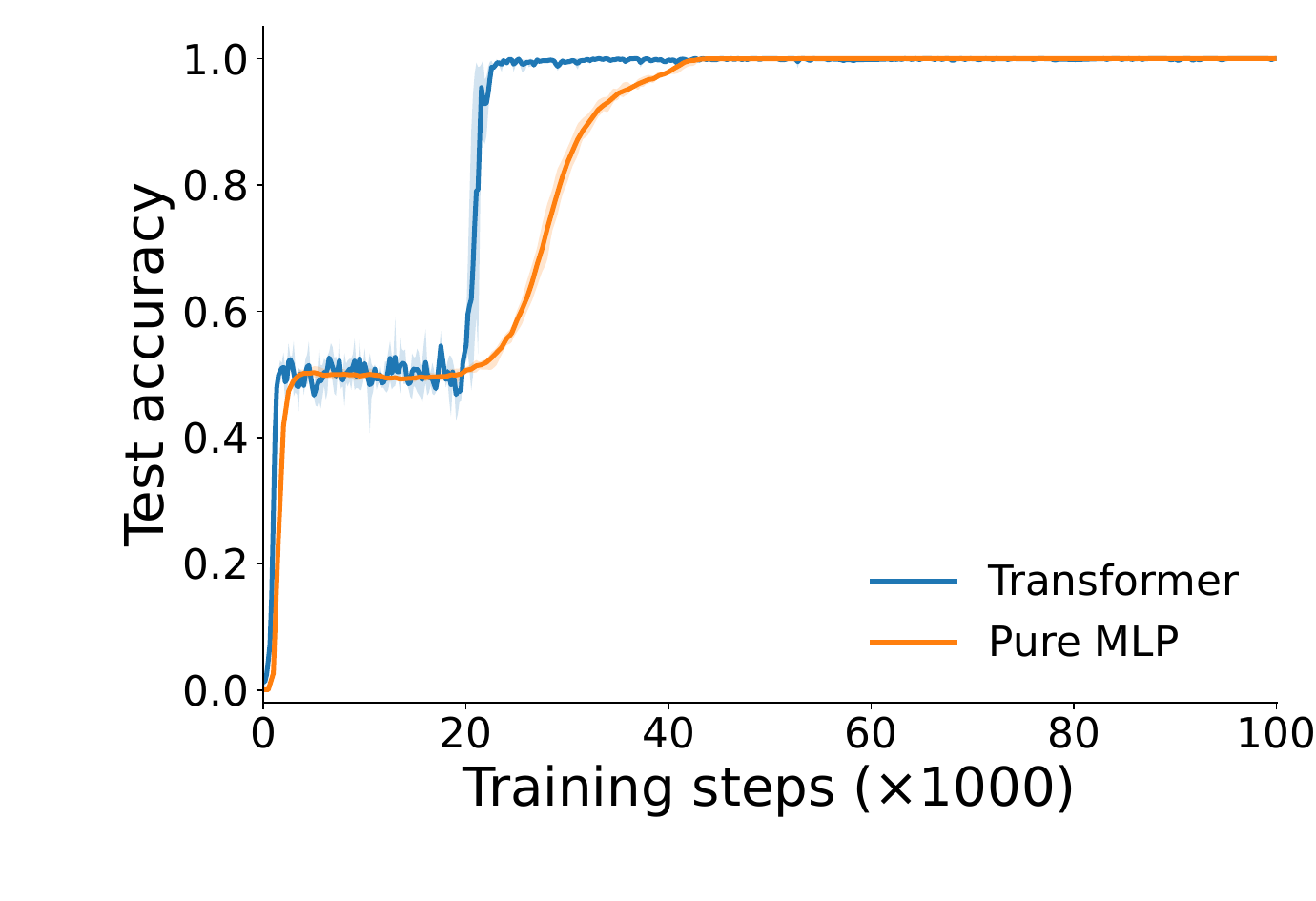}
        \caption{Training dynamics}
        \label{fig:D}
    \end{subfigure}
\caption{\textbf{Architectural inductive biases do not determine the emergence of symbolic mechanisms.}
      \textbf{(a)}~Flip rate at $V{=}10\text{k}$ for Transformers with specific inductive biases removed, and for MLP models.
      Error bars show the standard deviation across 3 seeds.
      \textbf{(b)}~Flip rate decreases with role-token bias in the same way for the pure MLP and the Transformer.
      \textbf{(c)}~MLP models show the same dependence of out-of-distribution (reversed-role) accuracy on vocabulary size and role-token bias as transformers: with high bias and a small vocabulary they fail the reversed test, and reversed accuracy recovers as vocabulary grows.
      \textbf{(d)}~Test accuracy over training.}
    \label{fig:main}
\end{figure}

We next asked whether symbolic mechanisms depend on the inductive biases of the transformer architecture. We tested this by evaluating three modified transformers, each ablating a specific architectural feature that seems relevant to symbolic computation, described below.


\subsection{Modified Transformers}

There are several aspects of the transformer architecture that seem relevant to its ability to implement symbolic computation, including  1) the presence of discrete embeddings with shared parameters, potentially enabling a form of compositionality, 2) the use of distinct embeddings for keys, queries, and values, potentially enabling a form of indirection (i.e., the use of one variable to refer to another), and 3) the use of inner products in the attention operation, which are well suited to encoding abstract relations \citep{webb2024relational,altabaa2024approximation}. We evaluated three architectural modifications against a standard transformer control, all trained at $V=10\text{k}$, to test whether specific inductive biases of the transformer architecture are necessary for the emergence of symbolic mechanisms (Figure~\ref{fig:A}). The first modified architecture removed weight sharing across sequence positions (No pos. weight sharing), ablating the shared parameters that may enable compositional representations. The second constrained the query, key, and value projection matrices to be identical (Shared Q/K/V), removing the separation that may encourage transformers to implement indirection. The third replaced dot-product attention with a learned scalar attention score computed by a small MLP over concatenated key and query representations (MLP attention), removing an inductive bias that may be helpful for learning abstract relations.

Figure~\ref{fig:A} shows the flip rate for our concept intervention at the third token position, measured at the final training checkpoint. Flip rates were high across all of the architectural variants: the control Transformer, the variant without position-specific weights, and the variant with shared Q/K/V all reached $100\%$, and the MLP attention variant reached about $90\%$ on average, with more variability across runs. None of these specific Transformer features is therefore necessary for symbolic mechanisms to emerge. This raises the question: do we need attention at all? 

\subsection{MLP Models}

To test whether symbolic mechanisms require attention at all, we trained a plain MLP with no attention. The MLP received as input the concatenation of all token embeddings in the sequence. Residual connections were included between layers, as in the Transformer. The network had 4 hidden layers, and the output layer projected from $d_\mathrm{model}$ to the vocabulary using only the slice of the hidden state corresponding to the final sequence position (Appendix \ref{app:details_models}). This MLP reached a flip rate of $100\%$ in all five runs (Figure~\ref{fig:A}), indicating that symbolic mechanisms can emerge in a model without attention.

\paragraph{Concept intervention results.}
We then asked whether symbolic mechanisms in the MLP depend on data diversity in the same way as they do in the Transformer. Figure~\ref{fig:B} shows flip rate as a function of role-token bias for both models, using the manipulation described in Section \ref{sec:diversity}. Flip rate decreased monotonically with bias for both the Transformer and the MLP, indicating that the emergence of symbolic mechanisms is driven by data diversity for both Transformers and MLPs. 

\paragraph{Out-of-distribution generalization.}
MLP models also showed a qualitatively similar pattern of out-of-distribution generalization (Figure~\ref{fig:C}). At $V=100$, accuracy on reversed-role prompts dropped as role-token bias increased, whereas at $V=10\text{k}$, generalization was perfect across all bias levels. These results indicate that out-of-distribution generalization was driven by data diversity in MLPs just as it was in Transformers.

\paragraph{Training efficiency.}
Finally, we investigated whether transformers and MLPs show differences in the efficiency with which they developed symbolic mechanisms over the course of training. Figure~\ref{fig:D} shows accuracy over training for transformer and MLP models at $V=10\text{k}$. The transformer reached ceiling after roughly 20k steps and the MLP after roughly 40k. This suggests that, while data diversity is the main driver of symbolic computation, the transformer offers a modest training efficiency advantage, accelerating convergence without changing the nature of the final solution.

\section{Data Diversity Also Drives the Emergence of Entity-Attribute Binding}
\label{sec:binding}

To test whether the effect of data diversity extends beyond the ABA/ABB task, we introduced a second task that required \emph{entity--attribute binding}. Each input contains two entity--attribute pairs followed by a query entity, $e_1\,a_1\,e_2\,a_2\,e_q$, and the model must output the attribute bound to the queried entity: $a_1$ if $e_q = e_1$, and $a_2$ if $e_q = e_2$. For example, the sequence \texttt{Alice friendly Bob tall Bob} has the answer \texttt{tall}. In the experiments, entities and attributes are arbitrary tokens, and each sequence uses a new pairing, so the answer cannot be predicted from token identity alone. That is, the model must link each entity to the attribute that follows it in-context.
The task is built around the \emph{binding ID mechanism} identified by \citet{feng2024binding}, in which a language model binds an entity to its attribute by giving both the same abstract identifier, independent of token identity. Here, this means that $e_1$ and $a_1$ would be bound (via vector addition) to one identifier and $e_2$ and $a_2$ would be bound to another. The model answers the query by first looking up the identifier of $e_q$ and then retrieving the attribute that's bound to this identifier. 
As in Section~\ref{sec:solution}, we performed causal mediation analysis and representational similarity analysis to confirm that the binding task also used invariant, symbol-like binding-ID representations rather than token-specific ones; full results are provided in Appendix~\ref{app:binding_mechanism}.

\begin{figure}[t]
    \centering
    \begin{subfigure}[t]{0.325\linewidth}
        \centering
        \includegraphics[width=\linewidth]{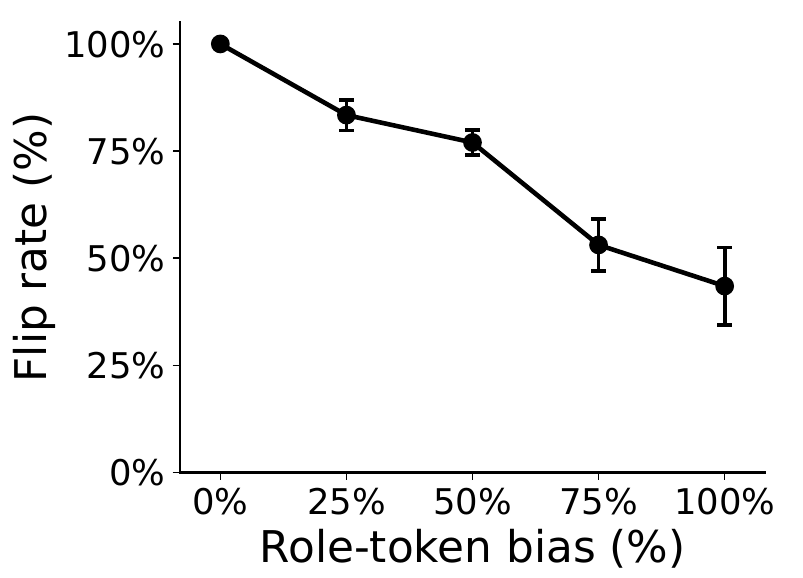}
        \caption{Flip rate vs.\ role-token bias}
        \label{fig:binding_bias}
    \end{subfigure}
    \hfill
    \begin{subfigure}[t]{0.325\linewidth}
        \centering
        \includegraphics[width=\linewidth]{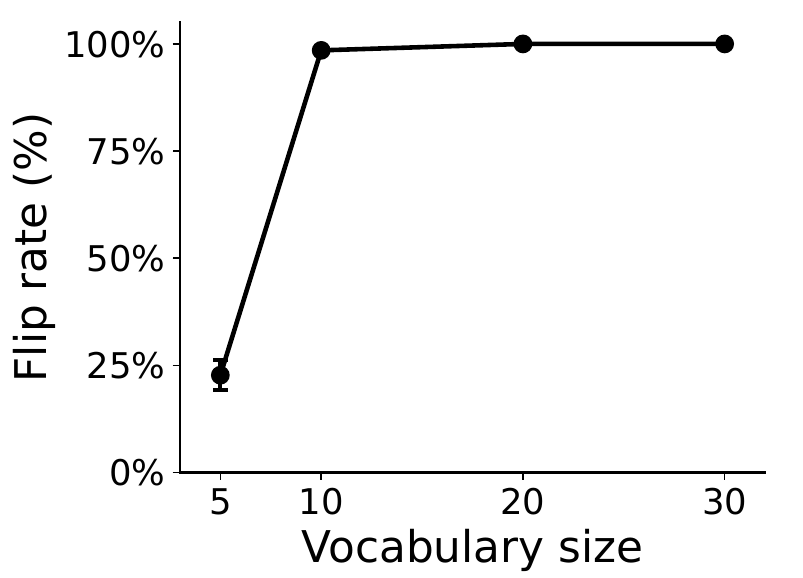}
        \caption{Flip rate vs.\ vocabulary size}
        \label{fig:binding_vocab}
    \end{subfigure}
    \hfill
    \begin{subfigure}[t]{0.325\linewidth}
        \centering
        \includegraphics[width=\linewidth]{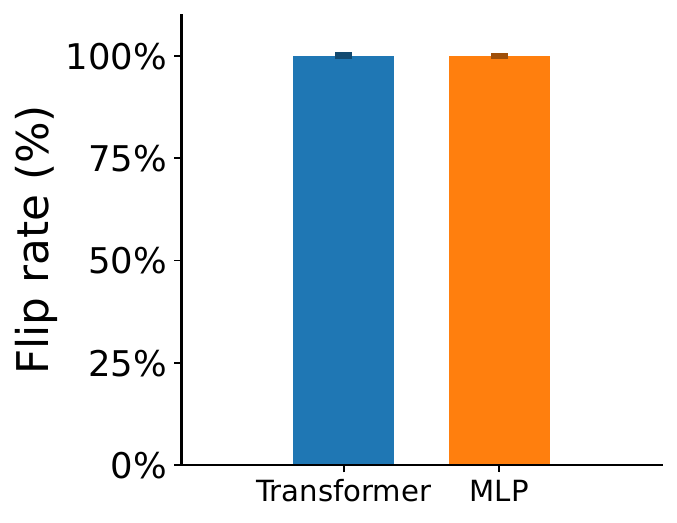}
        \caption{Transformer vs.\ MLP}
        \label{fig:binding_arch}
    \end{subfigure}
    \caption{\textbf{Data diversity drives the emergence of symbolic binding mechanisms.}
      \textbf{(a)}~Increasing role-token bias results in decreased binding-ID flip rates.
      \textbf{(b)}~The binding-ID flip rate increases with vocabulary size.
      \textbf{(c)}~Both MLPs and Transformers display high flip rates, indicating that both have developed symbolic binding mechanisms.}
    \label{fig:binding_main}
\end{figure}

Mirroring the role-token bias results in Section~\ref{sec:diversity}, increasing role-token bias degraded the binding-ID flip rate, from $100\%$ at $0\%$ bias to $44\%$ at $100\%$ bias (Figure~\ref{fig:binding_bias}). Consistent with the vocabulary-size results in Section~\ref{sec:diversity}, the flip rate also depended strongly on vocabulary size, with a flip rate of $23\%$ at $V{=}5$, and flip rates at or near ceiling for all larger vocabularies (Figure~\ref{fig:binding_vocab}). 
As in Section~\ref{sec:architecture}, this effect did not depend on the architectural inductive biases of the Transformer. Both a standard Transformer and an MLP displayed a binding-ID flip rate of $100\%$ (Figure~\ref{fig:binding_arch}). Together, these results indicate that the effect of data diversity on the emergence of symbolic mechanisms extends beyond the ABA/ABB rule task to a qualitatively different abstract reasoning task.

\section{Related Work}

\paragraph{Abstract reasoning in neural networks.}
A central debate in cognitive science and AI concerns whether neural networks perform genuine structured abstraction or simply exploit surface-level statistical patterns ~\citep{marcus2001algebraic,
mccoy2024embers, lewis2024evaluating, griffiths2025whither}.
\citet{webb2023emergent} demonstrated emergent analogical reasoning in LLMs, though its mechanisms remained unclear.
Our work builds on \citet{yang2025emergentsymbolicmechanismssupport}, by asking what
factors drive the emergence of a symbolic circuit for abstract reasoning in LLMs.

\paragraph{Mechanistic interpretability and in-context learning mechanisms.}
Our methods draw on a toolkit for reverse-engineering
neural network computations.
Causal mediation analysis~\citep{pearl2022direct, meng2022locating,
wang2022interpretability, todd2024function} localizes computations to components; we use it to
identify symbol abstraction heads and characterize their causal role.
Representational similarity analysis~\citep{kriegeskorte2008representational} complements this by
distinguishing abstract from token-specific encodings.
Our concept intervention method extends the activation patching paradigm
of \citet{meng2022locating} and the function vector framework of
\citet{todd2024function}.

\paragraph{Relational inductive biases and inner products.}
A complementary line of work asks how architectural design shapes
learned representations.
The relational bottleneck hypothesis~\citep{webb2024relational} proposes
that constraining networks to process relational rather than object-level
information encourages symbol-like representations. Abstractors~\citep{altabaa2024abstractors}
implement this via relational cross-attention.
Inner product spaces are well-suited to encoding abstract relations~\citep{altabaa2024approximation},
suggesting that the attention mechanism may be a useful inductive bias for symbolic computation.
Our finding that MLPs developed the same signatures qualifies this: inner products
may help (especially in terms of training efficiency) but are not necessary given sufficient data diversity.

\paragraph{Data and curriculum effects.}
A growing body of evidence shows that training data distributions shape
learned representations and strategies.
\citet{chan2022data} showed that in-context learning emerges from distributional properties like
burstiness and rare-class frequency. \citet{uselis2025does}
similarly found that compositional generalization in vision models is driven by data diversity, with greater
combinatorial coverage inducing a linearly factored representation.
Data diversity has also been linked to compositional generalization~\citep{hupkes2020compositionality}. We extend this to the mechanistic level: data diversity shapes not only behavioral generalization but the qualitative nature of the underlying mechanisms, independently of architecture.

\paragraph{Training dynamics.}
Language models often acquire different abilities at different points in training, sometimes abruptly. In-context learning emerges together with induction heads in a sudden phase change~\citep{olsson2022context,nallamala2024mechanistic}, and in-context learning develops in discrete stages~\citep{hoogland2024loss}. Generalization can also follow memorization after long training \citep{li2025tracing}, as in grokking~\citep{power2022grokking,nanda2023progress}. Our low-diversity models showed a similar ordering, first generalizing in-distribution via a statistical solution, and then developing symbolic mechanisms only after a long delay. How such phases relate to the emergence of specific symbolic mechanisms remains to be tested.

\section{Conclusion}

What determines whether a neural network develops symbolic mechanisms? We found that data diversity consistently promoted symbolic mechanisms across four axes, while architectural inductive biases were not necessary for their emergence. These findings extend to a second abstract reasoning task, suggesting that data diversity plays a central role in determining when and how symbolic mechanisms emerge.

\paragraph{Limitations and Future Work.}
The solution learned by our models partially reproduced the symbolic mechanisms observed in pretrained LLMs. Our models developed symbol abstraction heads, which LLMs use to \emph{infer} abstract rules from in-context examples, but not the symbolic induction heads and retrieval heads that LLMs use to \emph{apply} these rules to new examples. We hypothesize that greater rule diversity (in addition to token diversity) and rule uncertainty may be needed to induce these additional mechanisms, but this remains to be tested. Rather than attempting to reproduce the full suite of mechanisms employed by large-scale models, we chose symbol abstraction heads as a testbed for the factors that drive emergent symbolic computation. Two further limitations concern training dynamics and architecture. We trained small models from scratch on controlled tasks and found that low-diversity models first learned a statistical solution and developed symbolic mechanisms only after a long delay. How such dynamics unfold in large models trained under much more complex regimes, with mixed data, long training, and post-training, is unclear. Our concept interventions also showed that MLPs reached the same flip rates as Transformers, but how symbolic mechanisms are actually implemented in an MLP remains an open question.


\begin{ack}
This work was supported by the Courtois Chair in Fundamental Research - V (Neuroscience) at the Université de Montréal awarded to T.W. We are grateful to Mila (\href{https://mila.quebec}{mila.quebec}) and Princeton University for the compute resources, software, and technical help that made our experiments possible.
\end{ack}


\bibliographystyle{plainnat}
\bibliography{references}

@inproceedings{yang2025emergentsymbolicmechanismssupport,
  title={Emergent Symbolic Mechanisms Support Abstract Reasoning in Large Language Models},
  author={Yang, Yukang and Campbell, Declan and Huang, Kaixuan and Wang, Mengdi and Cohen, Jonathan and Webb, Taylor},
  booktitle={International Conference on Machine Learning (ICML)},
  year={2025}
}

@article{mccoy2024embers,
  title={Embers of autoregression show how large language models are shaped by the problem they are trained to solve},
  author={McCoy, R Thomas and Yao, Shunyu and Friedman, Dan and Hardy, Mathew D and Griffiths, Thomas L},
  journal={Proceedings of the National Academy of Sciences},
  volume={121},
  number={41},
  pages={e2322420121},
  year={2024},
  publisher={National Academy of Sciences}
}

@article{lewis2024evaluating,
  title={Evaluating the robustness of analogical reasoning in large language models},
  author={Lewis, Martha and Mitchell, Melanie},
  journal={arXiv preprint arXiv:2411.14215},
  year={2024}
}

@article{webb2023emergent,
  title={Emergent analogical reasoning in large language models},
  author={Webb, Taylor and Holyoak, Keith J and Lu, Hongjing},
  journal={Nature Human Behaviour},
  volume={7},
  number={9},
  pages={1526--1541},
  year={2023},
  publisher={Nature Publishing Group UK, London}
}

@article{griffiths2025whither,
  title   = {Whither Symbols in the Era of Advanced Neural Networks?},
  author  = {Griffiths, Thomas L. and Lake, Brenden M. and McCoy, R. Thomas and Pavlick, Ellie and Webb, Taylor W.},
  journal = {Trends in Cognitive Sciences},
  year    = {2026},
  doi     = {10.1016/j.tics.2026.02.003}
}

@book{marcus2001algebraic,
  title={The algebraic mind: Integrating connectionism and cognitive science},
  author={Marcus, Gary F},
  year={2001},
  publisher={The MIT Press}
}

@inproceedings{altabaa2024abstractors,
  title={Abstractors and relational cross-attention: An inductive bias for explicit relational reasoning in transformers},
  author={Altabaa, Awni and Webb, Taylor and Cohen, Jonathan and Lafferty, John},
  booktitle={International Conference on Learning Representations},
  volume={2024},
  pages={24597--24613},
  year={2024}
}

@incollection{pearl2022direct,
  title={Direct and indirect effects},
  author={Pearl, Judea},
  booktitle={Probabilistic and causal inference: the works of Judea Pearl},
  pages={373--392},
  year={2022},
  publisher={ACM}
}

@article{webb2024relational,
  title={The relational bottleneck as an inductive bias for efficient abstraction},
  author={Webb, Taylor W and Frankland, Steven M and Altabaa, Awni and Segert, Simon and Krishnamurthy, Kamesh and Campbell, Declan and Russin, Jacob and Giallanza, Tyler and O’Reilly, Randall and Lafferty, John and others},
  journal={Trends in Cognitive Sciences},
  volume={28},
  number={9},
  pages={829--843},
  year={2024},
  publisher={Elsevier}
}

@article{altabaa2024approximation,
  title={Approximation of relation functions and attention mechanisms},
  author={Altabaa, Awni and Lafferty, John},
  journal={arXiv preprint arXiv:2402.08856},
  year={2024}
}

@article{uselis2025does,
  title={Does data scaling lead to visual compositional generalization?},
  author={Uselis, Arnas and Dittadi, Andrea and Oh, Seong Joon},
  journal={arXiv preprint arXiv:2507.07102},
  year={2025}
}

@article{hupkes2020compositionality,
  title={Compositionality decomposed: How do neural networks generalise?},
  author={Hupkes, Dieuwke and Dankers, Verna and Mul, Mathijs and Bruni, Elia},
  journal={Journal of Artificial Intelligence Research},
  volume={67},
  pages={757--795},
  year={2020}
}

@article{olsson2022context,
  title={In-context learning and induction heads},
  author={Olsson, Catherine and Elhage, Nelson and Nanda, Neel and Joseph, Nicholas and DasSarma, Nova and Henighan, Tom and Mann, Ben and Askell, Amanda and Bai, Yuntao and Chen, Anna and others},
  journal={arXiv preprint arXiv:2209.11895},
  year={2022}
}

@inproceedings{feng2024binding,
  title     = {How Do Language Models Bind Entities in Context?},
  author    = {Feng, Jiahai and Steinhardt, Jacob},
  booktitle = {International Conference on Learning Representations (ICLR)},
  year      = {2024},
  url       = {https://arxiv.org/abs/2310.17191}
}

@article{wang2022interpretability,
  title={Interpretability in the wild: a circuit for indirect object identification in GPT-2 small},
  author={Wang, Kevin and Variengien, Alexandre and Conmy, Arthur and Shlegeris, Buck and Steinhardt, Jacob},
  journal={arXiv preprint arXiv:2211.00593},
  year={2022}
}

@article{meng2022locating,
  title={Locating and editing factual associations in gpt},
  author={Meng, Kevin and Bau, David and Andonian, Alex and Belinkov, Yonatan},
  journal={Advances in neural information processing systems},
  volume={35},
  pages={17359--17372},
  year={2022}
}

@inproceedings{todd2024function,
  title={Function vectors in large language models},
  author={Todd, Eric and Li, Millicent and Sen Sharma, Arnab and Mueller, Aaron and Wallace, Byron and Bau, David},
  booktitle={International conference on learning representations},
  volume={2024},
  pages={17282--17333},
  year={2024}
}

@article{kriegeskorte2008representational,
  title={Representational similarity analysis-connecting the branches of systems neuroscience},
  author={Kriegeskorte, Nikolaus and Mur, Marieke and Bandettini, Peter A},
  journal={Frontiers in systems neuroscience},
  volume={2},
  pages={249},
  year={2008},
  publisher={Frontiers}
}

@inproceedings{nallamala2024mechanistic,
  title={The mechanistic basis of data dependence and abrupt learning in an in-context classification task},
  author={Nallamala, Gautam Reddy},
  booktitle={International conference on learning representations},
  volume={2024},
  pages={19226--19239},
  year={2024}
}

@article{hoogland2024loss,
  title   = {Loss Landscape Degeneracy and Stagewise Development in Transformers},
  author  = {Hoogland, Jesse and Wang, George and Farrugia-Roberts, Matthew and Carroll, Liam and Wei, Susan and Murfet, Daniel},
  journal = {Transactions on Machine Learning Research},
  year    = {2025},
  note    = {First posted as ``The Developmental Landscape of In-Context Learning''}
}

@article{power2022grokking,
  title={Grokking: Generalization beyond overfitting on small algorithmic datasets},
  author={Power, Alethea and Burda, Yuri and Edwards, Harri and Babuschkin, Igor and Misra, Vedant},
  journal={arXiv preprint arXiv:2201.02177},
  year={2022}
}

@article{nanda2023progress,
  title={Progress measures for grokking via mechanistic interpretability},
  author={Nanda, Neel and Chan, Lawrence and Lieberum, Tom and Smith, Jess and Steinhardt, Jacob},
  journal={arXiv preprint arXiv:2301.05217},
  year={2023}
}

@inproceedings{li2025tracing,
  title     = {Tracing the Representation Geometry of Language Models from Pretraining to Post-training},
  author    = {Li, Melody Zixuan and Agrawal, Kumar Krishna and Ghosh, Arna and Teru, Komal Kumar and Santoro, Adam and Lajoie, Guillaume and Richards, Blake A.},
  booktitle = {Advances in Neural Information Processing Systems (NeurIPS)},
  year      = {2025},
  note      = {arXiv:2509.23024}
}

@article{radford2019language,
  title={Language models are unsupervised multitask learners},
  author={Radford, Alec and Wu, Jeffrey and Child, Rewon and Luan, David and Amodei, Dario and Sutskever, Ilya and others},
  journal={OpenAI blog},
  volume={1},
  number={8},
  pages={9},
  year={2019}
}

@inproceedings{chan2022data,
  title     = {Data distributional properties drive emergent in-context learning in transformers},
  author    = {Chan, Stephanie and Santoro, Adam and Lampinen, Andrew and Wang, Jane and Singh, Aaditya and Richemond, Pierre and McClelland, James and Hill, Felix},
  booktitle = {Advances in Neural Information Processing Systems},
  volume    = {35},
  pages     = {18878--18891},
  year      = {2022}
}

\appendix


\newpage
\section{Experimental Details}
\label{app:details}

\subsection{Task and data}
\label{app:details_task}

\paragraph{ABA/ABB.} Each prompt contains one in-context example that follows a rule, followed by a query that the model must complete under the same rule. The input is five tokens, A1 B1 A1 A2 B2 (ABA) or A1 B1 B1 A2 B2 (ABB), and the model predicts the sixth token (A2 for ABA, B2 for ABB). There is no separator or start token. The four tokens A1, B1, A2, B2 are distinct and are drawn without replacement from the vocabulary \texttt{tok\_1}, \dots, \texttt{tok\_$V$}, freshly for every prompt, so surface identity carries no information about the rule. The two rules are equally likely. 

\paragraph{Binding.} A prompt is $[e_1, a_1, e_2, a_2, q]$, where $e_1\neq e_2$ are entities, $a_1\neq a_2$ are attributes, $q$ is one of the two entities (chosen with probability $1/2$), and the target is the attribute paired with $q$. The vocabulary has about 20{,}000 tokens.

\paragraph{Dataset sizes.} Table~\ref{tab:details_data} lists the training and evaluation sets used in each experiment. 

\begin{table}[h]
\centering\small
\caption{Datasets. ``Prompts'' counts distinct training prompts.}
\label{tab:details_data}
\begin{tabular}{llrr}
\toprule
Experiment & Vocabulary & Train prompts & Validation prompts \\
\midrule
Vocabulary size (ABA/ABB) & 50--10{,}000 & 2{,}000{,}000 & 2{,}000 \\
Number of training samples & 10{,}000 & 200--200{,}000 & 500 \\
Number of rules & 100 & 2{,}000{,}000 & 5{,}000 \\
Role-token bias & 100--20k & 5{,}000{,}000 & 5{,}000 \\
Architecture variants & 10{,}000 & 5{,}000{,}000 & 10{,}000 \\
Binding (role-token bias) & 20{,}000 & 5{,}000{,}000 & 20{,}000 \\
\bottomrule
\end{tabular}
\end{table}
The role-token bias datasets additionally contain a standard test set (no role-exclusive tokens) and a reversed-role test set of 5{,}000 prompts each.

\subsection{Models}
\label{app:details_models}

All models are trained from scratch. Tokens are mapped to ids by a fixed lookup table with a few special tokens (pad, unk, bos, eos, sep).

\paragraph{Transformer.} GPT-2 as implemented in HuggingFace \texttt{transformers} (4.56.2) with the GPT-2-small configuration: 12 layers, 12 attention heads, hidden size 768, MLP width 3072, GELU, pre-LayerNorm, learned absolute position embeddings, tied input and output embeddings, and dropout 0.1 on embeddings, attention and residual connections. Weights use the default GPT-2 initialization. The model has about 86M parameters at $V=100$ and 94M at $V=10{,}000$.

\paragraph{Architecture variants (Section~\ref{sec:architecture}).} All variants keep the Transformer above except for the one change listed.
\begin{itemize}\setlength\itemsep{0pt}
\item \emph{No positional weight sharing.} Every input position has its own MLP weights (\texttt{c\_fc}, \texttt{c\_proj}) in every block; attention weights stay shared across positions.
\item \emph{Shared Q/K/V.} A single projection $W_s\in\mathbb{R}^{768\times768}$ per layer produces queries, keys and values ($q=k=v=W_s h$), which makes the attention scores symmetric.
\item \emph{MLP attention.} The dot-product score of each head is replaced by a two-layer network applied to every (query, key) pair, $s(q,k)=W_2\,\mathrm{GELU}(W_1[q;k]+b_1)+b_2$, with hidden size equal to the head dimension (64). The softmax, causal mask, value and output projections are unchanged.
\item \emph{Pure MLP (no attention).} The input is padded to 16 positions. Token and learned position embeddings (dimension 64) are summed and the sequence is flattened into a 1024-dimensional vector. Four residual blocks follow, each LayerNorm, Linear $1024\!\to\!2048$, GELU, dropout 0.1, Linear $2048\!\to\!1024$; then a final LayerNorm, a reshape to $16\times64$, and a tied unembedding. There is no causal mask, so every position can see every other; the target token is never given as input, and the prediction is read out at the last context position. The model has about 17M parameters.
\end{itemize}

\paragraph{Binding models.} The Transformer is the same GPT-2-small configuration with a context length of 10. The MLP is the pure MLP above with 8 blocks instead of 4 (about 35M parameters).

\subsection{Training}
\label{app:details_training}

All models are trained with the HuggingFace \texttt{Trainer}: AdamW ($\beta_1=0.9$, $\beta_2=0.999$, $\epsilon=10^{-8}$) with no weight decay, gradient-norm clipping at 1.0, fp16 mixed precision, a linear warmup over the first 10\% of steps followed by cosine decay to zero, and per-device batch size 24 on one GPU. The peak learning rate is $10^{-4}$ in all experiments except the MLP-attention variant ($3\times10^{-4}$, two GPUs with per-device batch 24, i.e.\ an effective batch of 48). Dropout is 0.1.

\subsection{Intervention and flip rate}
\label{app:details_flip}

\paragraph{Role vectors.} For every layer $\ell$ we take the residual stream after block $\ell$ at the third token of the in-context example (where the ABA and ABB examples differ). The ABA vector $z_{\text{ABA}}^{\ell}$ is the mean of this activation over $2n$ ABA prompts, and $z_{\text{ABB}}^{\ell}$ is the mean over $2n$ ABB prompts. The ABA and ABB prompts are built from the same tokens (each of $n$ token pairs is used in both orders, and the in-context and query tokens are shared), so the two sets differ only in the rule. We use $n=500$ in all reported results, i.e.\ 1{,}000 prompts per rule.

\paragraph{Intervention.} On a test ABA prompt we add $z_{\text{ABB}}^{\ell}-z_{\text{ABA}}^{\ell}$ to the residual stream at that position, at every layer simultaneously (and the reverse for ABB prompts), this is equivalent to replacing the rule component $z_{\text{ABA}}$ by $z_{\text{ABB}}$ while keeping the token-specific component. All tokens are unchanged. For the pure MLP, which has no per-position streams, the vector is added to the flattened hidden vector at every block. For the 5-token rules, the vector is estimated and applied at the last token of the in-context example (position 4), where ABABA and ABABB first differ.

\paragraph{Flip rate.} A test prompt \emph{flips} if the model's prediction after the intervention exactly equals the correct answer under the opposite rule for that prompt. Test prompts use fresh tokens, 200 per rule for the reported results (400 prompts in total) with duplicates removed.  For the out of distribution bias experiments the flip rate is measured on reversed-role prompts, using role vectors estimated from role-respecting prompts of the same bias level.

\paragraph{Uncertainty.} Points are means over 3-5 seeds; error bars are standard deviations (rules, vocabulary size) or the min--max range (bias, architectures), as stated in each caption.

\subsection{Data-diversity manipulations}
\label{app:details_diversity}

\paragraph{Vocabulary size.} The number of distinct tokens in the training dataset, $V\in\{50,100,500,1000,2000,4000,8000,10000\}$.
\paragraph{Number of training samples.} The number of distinct training prompts, $N\in\{200, 1000, 5000, 20000, 50000, 100000, 200000\}$ at $V=10{,}000$. 

\paragraph{Number of rules.}For $R=10$ the rules are length-5 patterns: ABABA, ABABB, ABCBA, ABCBB, ABCCC, AAABA, AAABB, AABAA, AABAB, AABBA, at $V=100$.  The $R=2$ point is the plain ABA/ABB task at $V=100$.

\paragraph{Role-token bias.} A fraction $b\in\{0,0.25,0.5,0.75,1\}$ of the vocabulary is role-exclusive during training, split evenly into a set that only fills role A and a set that only fills role B; the remaining $(1-b)V$ tokens can fill either role. At test time the two role-exclusive sets are swapped (the reversed-role test) while the rule is unchanged. For the binding task the exclusive sets are entity-only and attribute-only tokens.

\subsection{Convergence definitions}
\label{app:details_convergence}

Accuracy is evaluated on the validation set every 250 steps and flip rate at every saved checkpoint. A quantity \emph{converges} at the first checkpoint at which its rolling mean over three checkpoints reaches the threshold and at least 90\% of the remaining checkpoints stay above 95\% of the threshold. Accuracy convergence uses a threshold of 95\% in the vocabulary-size and number-of-rules plots. Flip-rate convergence uses 90\%, and runs that never reach it are excluded from the ratio plots. The flip rate at accuracy convergence is read from the checkpoint nearest to the accuracy-convergence step.

\section{Decoder Experiments}
\label{app:decoder}

To test the account in Section~\ref{sec:rotation} that early layers carry abstract symbol identity while later layers carry the correct answer, we trained linear probes on the residual stream at the final query position of a model (vocabulary size 2000). The query always used the same two tokens, so token identity was constant across trials and only (i) which of the two tokens plays role A and (ii) which rule is active vary. Each trial has one in-context example with fresh tokens. We used 400 trials per rule. For each layer, we trained a logistic-regression probe (L2 penalty, $C=1$) on raw activations from one rule (ABA or ABB) to predict whether the first query token is the correct answer, then evaluated the same probe on the \emph{other} rule against two different labels:
\begin{itemize}
    \item \textbf{symbol generalization}: whether the first query token has the role the probe was trained to detect (role A for an ABA-trained probe, role B for an ABB-trained probe)
    \item \textbf{correct/incorrect generalization}: whether the first query token is the correct answer, which flips meaning between the two rules.
\end{itemize}
Within a single rule the two labels coincide, so only the cross-rule test separates them. Chance is 50\%. A probe that tracks role identity scores 100\% on the first label and 0\% on the second, because the correct token changes role between the rules; a probe that tracks the answer scores the reverse. We average over both training directions (ABA$\to$ABB and ABB$\to$ABA).

\begin{figure}[t]
    \centering
    \includegraphics[width=0.55\textwidth]{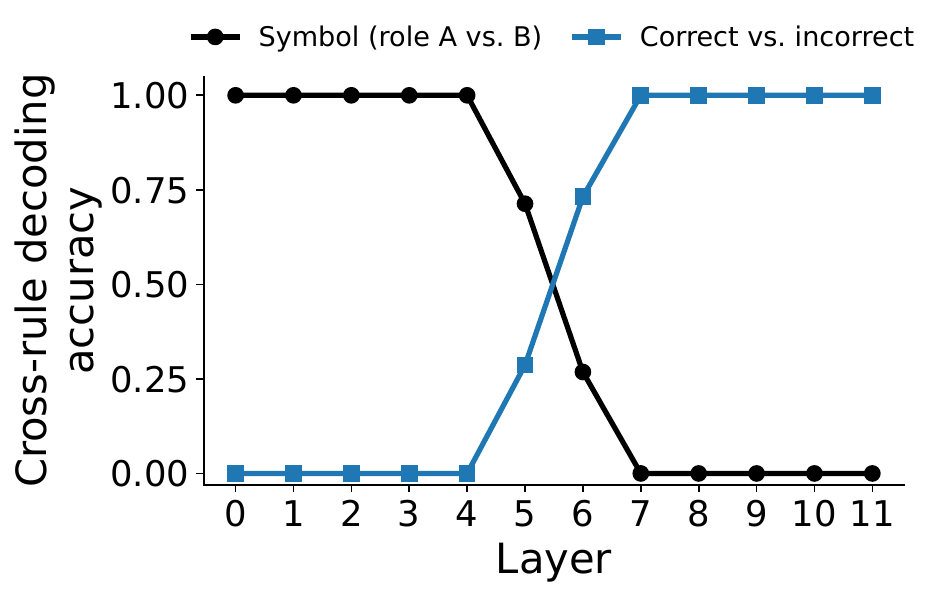}
    \caption{\textbf{Cross-rule decoding by layer.} Probes trained on one rule and tested on the other. Layers 0--4 decode symbol identity perfectly; their correct/incorrect score is 0\% because the same direction points to the wrong token under the new rule. Layers 7--11 show the reverse.}
    \label{fig:decoder}
\end{figure}

Symbol generalization was $100\%$ in layers 0--4 and fell to $71\%$ and $27\%$ in layers 5 and 6 and $0\%$ afterwards, while correct/incorrect generalization rose from $0\%$ (layers 0--4) through $29\%$ and $73\%$ (layers 5 and 6) to $100\%$ from layer 7 onward (Figure~\ref{fig:decoder}). Because the cross-rule probe is evaluated on a rule it was never trained on, this is evidence that early layers represent role identity abstractly and later layers represent the answer itself. 

\section{Training Dynamics Across Data Diversity Levels}
\label{app:diversity_dynamics}

\begin{figure}
    \centering
    \includegraphics[width=\textwidth]{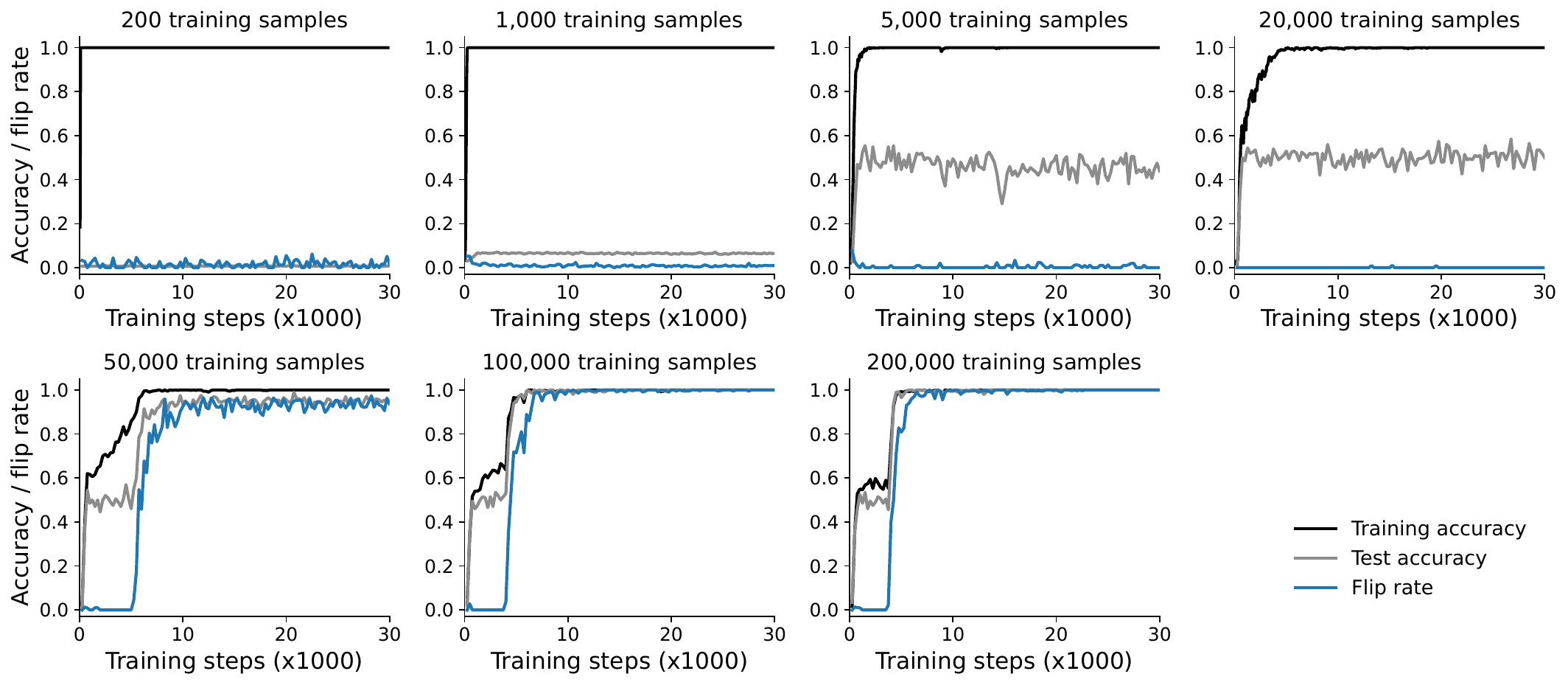}
    \caption{\textbf{Timecourses by number of training samples} (vocabulary size 10{,}000). Flip rate is measured on in-distribution prompts.}
    \label{fig:tc_ntrain}
\end{figure}

\begin{figure}
    \centering
    \includegraphics[width=\textwidth]{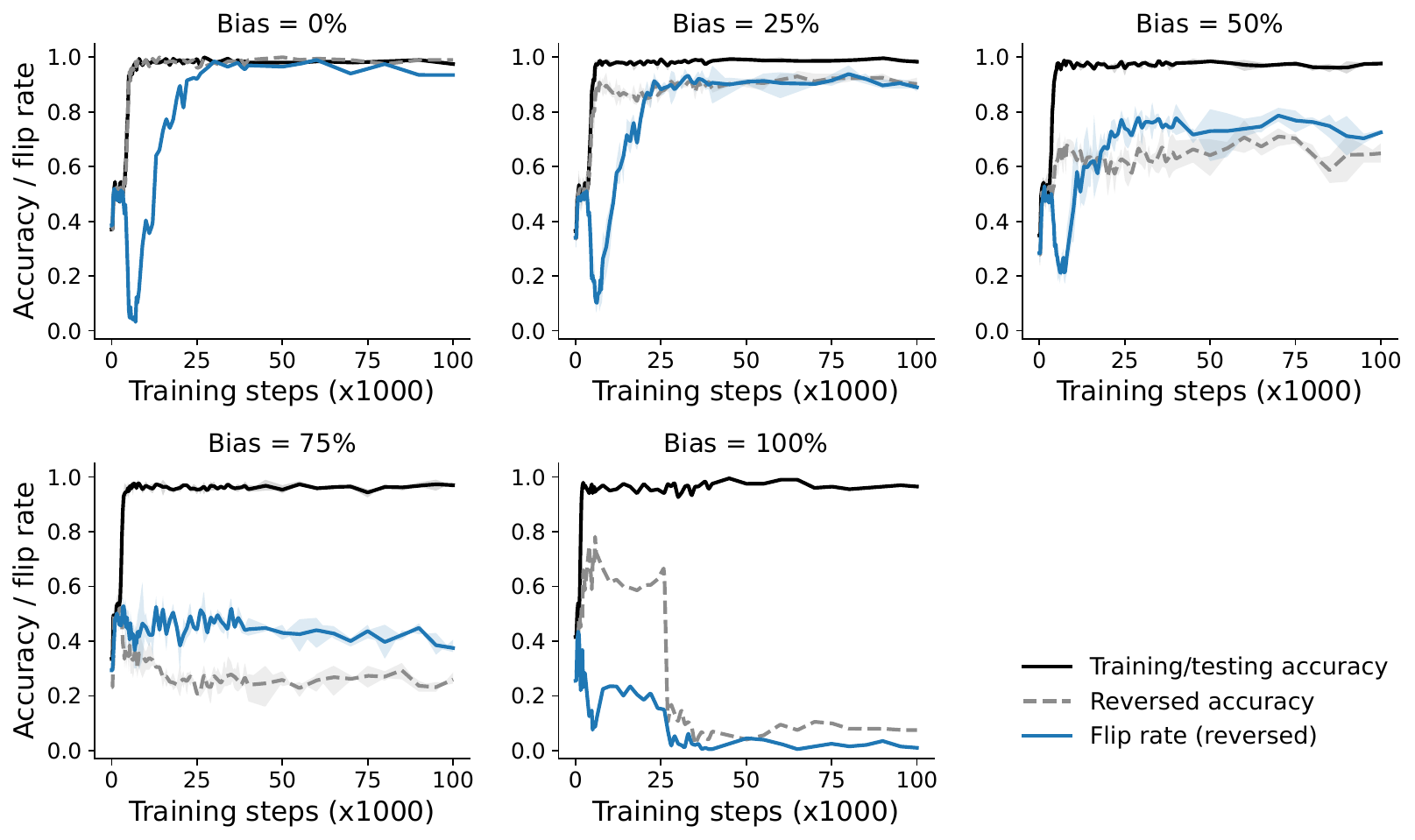}
    \caption{\textbf{Timecourses by role-token bias} (vocabulary size 100). Solid black: in-distribution accuracy. Dashed gray: accuracy on reversed-role (out-of-distribution) prompts. Blue: flip rate on reversed-role prompts. In-distribution accuracy is near ceiling at every bias level, while reversed accuracy and reversed flip rate fall as bias increases. } 
    \label{fig:tc_bias}
\end{figure}

\begin{figure}
    \centering
    \includegraphics[width=\textwidth]{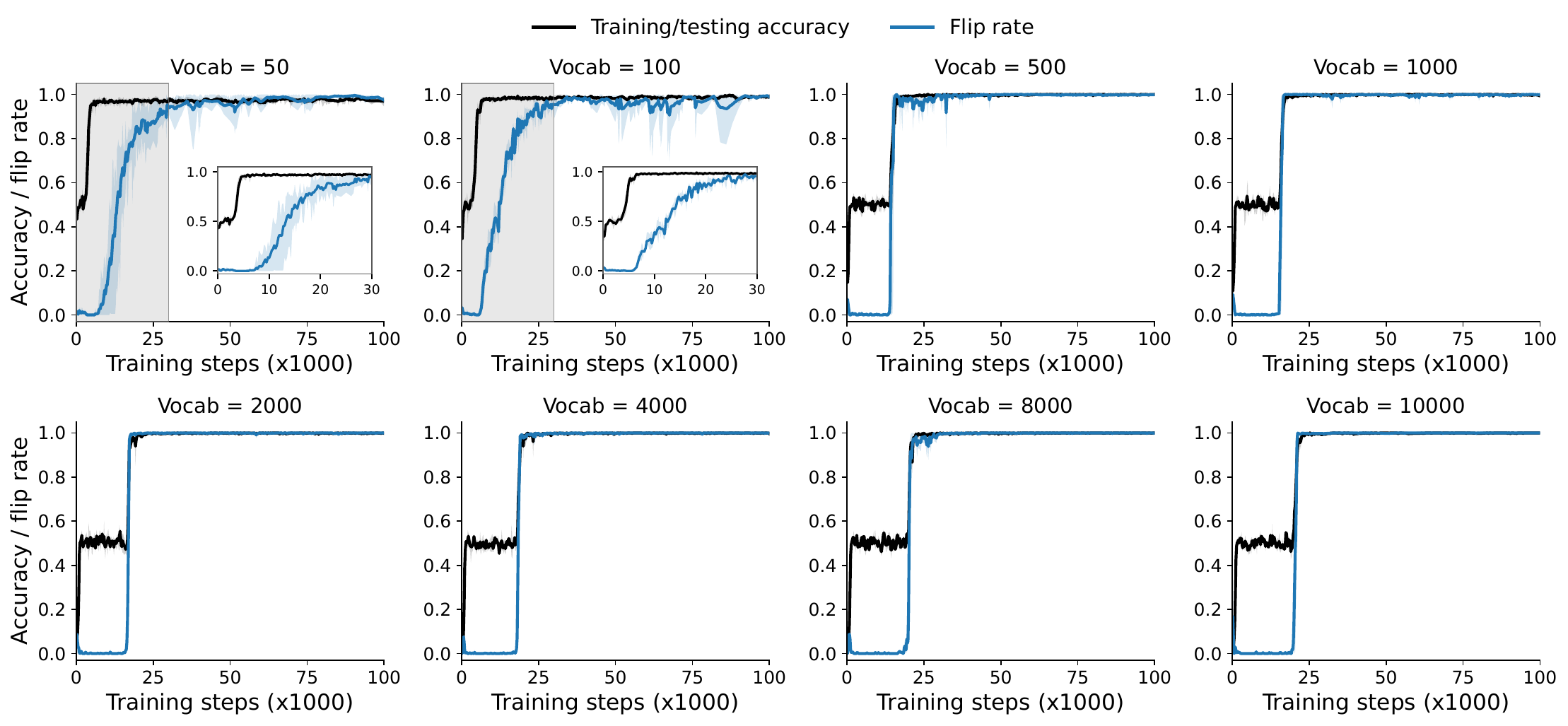}
    \caption{\textbf{Timecourses by vocabulary size}. Shaded boxes and insets zoom on the first 30{,}000 steps for the two smallest vocabularies. At small vocabularies ($V\le100$) flip rate rises well after accuracy has converged; from $V=500$ upward the two curves rise together.}
    \label{fig:tc_vocab}
\end{figure}

\begin{figure}
    \centering
    \includegraphics[width=0.7\textwidth]{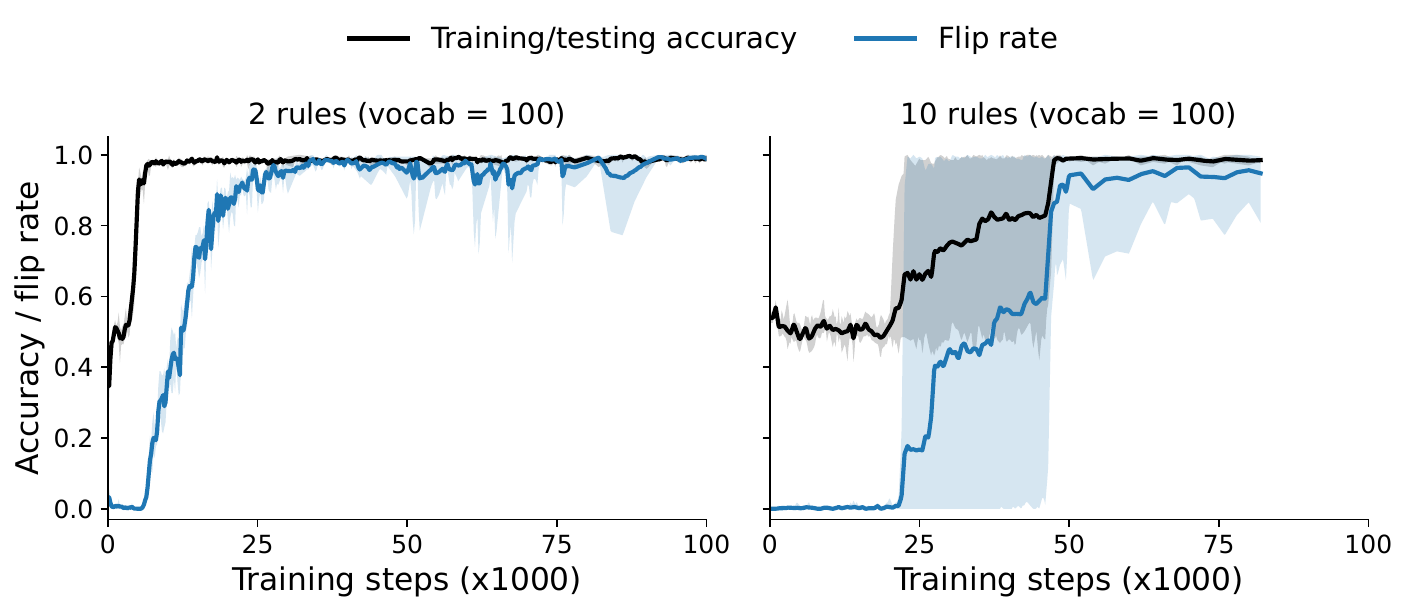}
    \caption{\textbf{Timecourses by number of rules} (vocabulary size 100). The 2-rule condition is the plain ABA/ABB task. The 10-rule condition uses ten length-5 rules.}
    \label{fig:tc_rules}
\end{figure}

\begin{figure}
    \centering
    \includegraphics[width=\textwidth]{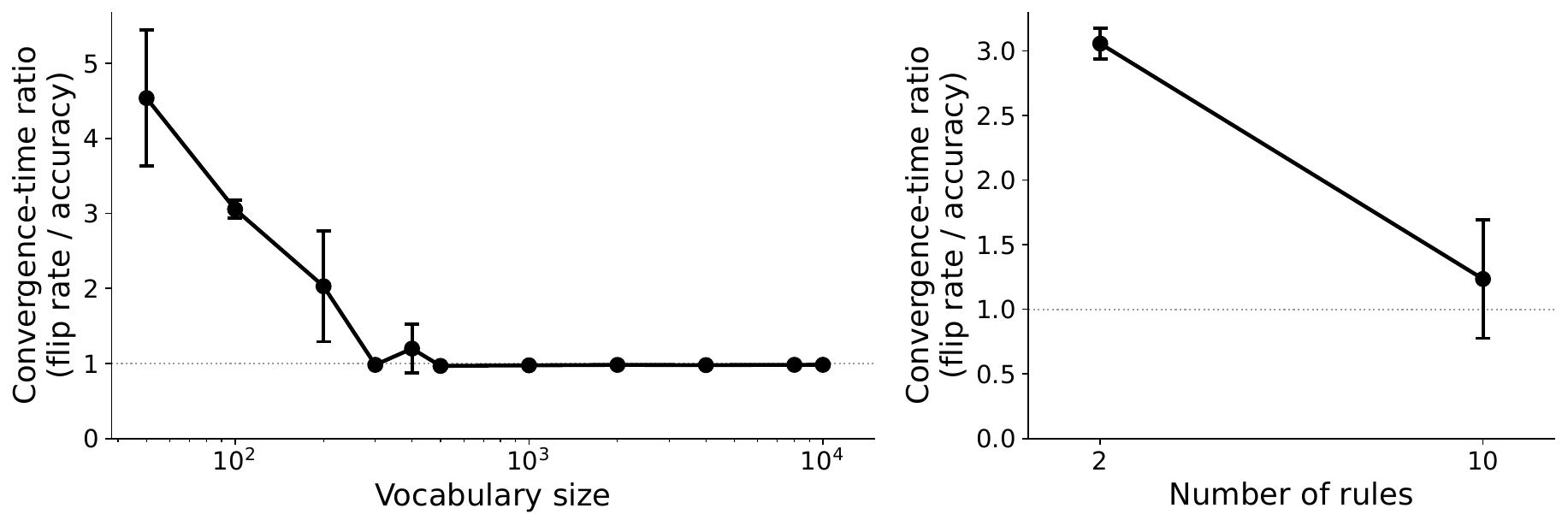}
    \caption{\textbf{Ratio of flip-rate to accuracy convergence step across diversity levels.} A ratio of 1 means the mechanism converges together with accuracy. Points are means over runs. The ratio is about 1 for vocabulary sizes $V\ge500$ and for the number-of-rules axis, and larger for the smallest vocabularies. } 
    \label{fig:conv_ratio}
\end{figure}

This section gives the training dynamics behind the data-diversity results in the main text. For each level of each diversity axis, we track task accuracy and the third-token flip rate at every saved checkpoint. We include all four data-diversity axes: number of training samples (Figure~\ref{fig:tc_ntrain}), role-token bias (Figure~\ref{fig:tc_bias}), vocabulary size (Figure~\ref{fig:tc_vocab}), and number of rules (Figure~\ref{fig:tc_rules}). Unless stated otherwise, accuracy is shown in black and flip rate in blue; lines are the mean over seeds and the shaded band spans the minimum and maximum over seeds. Accuracy is measured on freshly generated test prompts. Figure~\ref{fig:conv_ratio} summarizes the same data as the ratio of the flip-rate convergence step to the accuracy convergence step.

\label{app:timecourses}

\section{Effect of Model Size}
\label{app:model_size}

To test whether the delayed emergence of the symbolic mechanism depends on model capacity, we compared the default GPT-2-small model used throughout the paper (12 layers, 12 heads, hidden size 768; 85.9M parameters) with a much smaller model (2 layers, 2 heads, hidden size 128; 0.53M parameters) on the ABA/ABB task at vocabulary size 50.

\begin{figure}
    \centering
    \includegraphics[height=4.4cm]{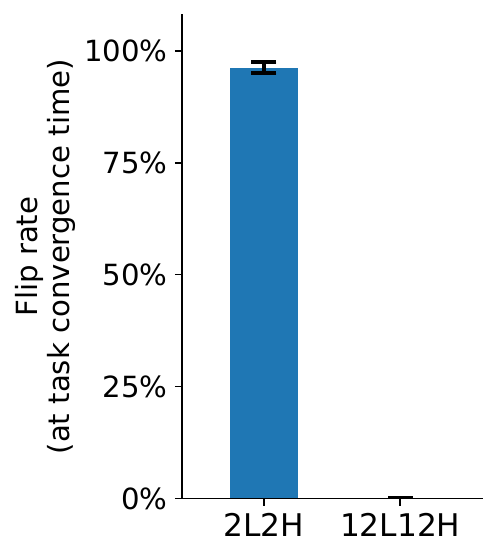}\hfill
    \includegraphics[height=4.4cm]{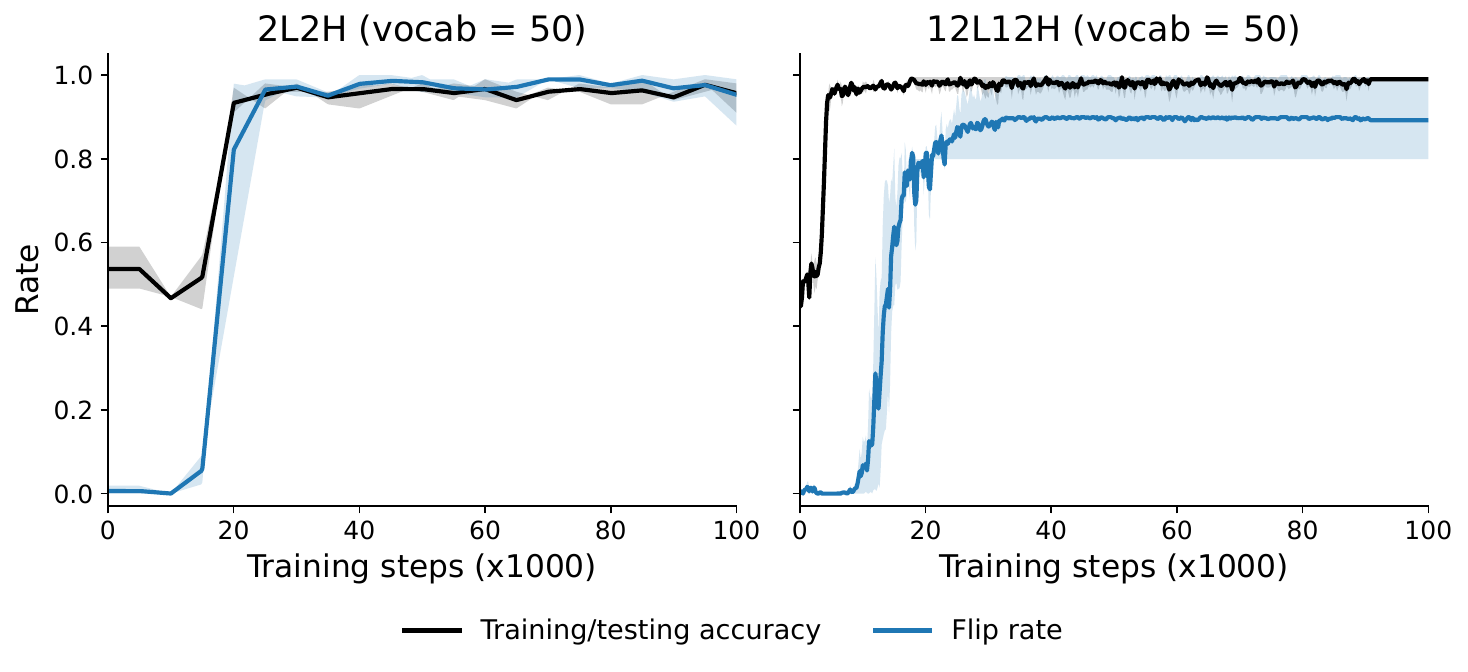}
    \caption{\textbf{Effect of model size on the emergence of the symbolic mechanism} (vocabulary size 50). \textbf{Left:} flip rate at the step where accuracy first converges, mean $\pm$ standard deviation over seeds. \textbf{Right:} training dynamics of the two models: accuracy and flip rate over training.}
    \label{fig:model_size}
\end{figure}

\paragraph{Smaller models skip the statistical phase.} At the step where accuracy converges, the 2L2H model already had a flip rate of $96.2\% \pm 1.3\%$, whereas the 12L12H model had a flip rate of $0\%$ (Figure~\ref{fig:model_size}). The 12L12H model reached 95\% accuracy much earlier, but its flip rate only rose about $6.5\times$ later; in the 2L2H model the two converge together. Thus the smaller model does not show the initial statistical phase at all: it appears to go directly to symbolic signatures, although it takes longer, in training steps, to get there than the base model takes to solve the task.

\paragraph{Capacity versus diversity.} We think this fits naturally alongside the vocabulary-size result. Reducing capacity, by using a smaller model, removes the option of a first-order statistical solution outright, so there is no memorization phase to pass through. Reducing diversity, by using a smaller vocabulary at fixed model size, merely makes the statistical solution viable for longer before the pressure toward the compressed, symbolic solution eventually wins out.

\section{Mechanistic Analysis of Entity-Attribute Binding}
\label{app:binding_mechanism}

We tested whether the entity-attribute binding task (Section~\ref{sec:binding}) is solved with binding IDs, mirroring the analyses used for the ABA/ABB task in Section~\ref{sec:solution}: an intervention, a causal mediation analysis, and a representational similarity analysis.

\subsection{Binding-ID intervention}
\label{app:binding_intervention}

Each input lists two entity-attribute pairs followed by a query entity, and the model must output the attribute bound to it. Under the binding-ID hypothesis, the model tags each pair with an ID (ID0 or ID1) and answers by matching the query's ID. We therefore intervene at the query position: for a query on the first entity we subtract the mean residual vector associated with ID0 and add the one associated with ID1 (and vice versa for the second entity). The two ID vectors are the mean residual-stream vector at the query position over many prompts in which the query is the first-pair entity (ID0) or the second-pair entity (ID1), estimated on one half of the entities and tested on the held-out half, so the test queries are novel tokens. A \emph{flip} means the model now outputs the attribute bound to the other entity.

\begin{figure}[t]
    \centering
    \includegraphics[width=\textwidth]{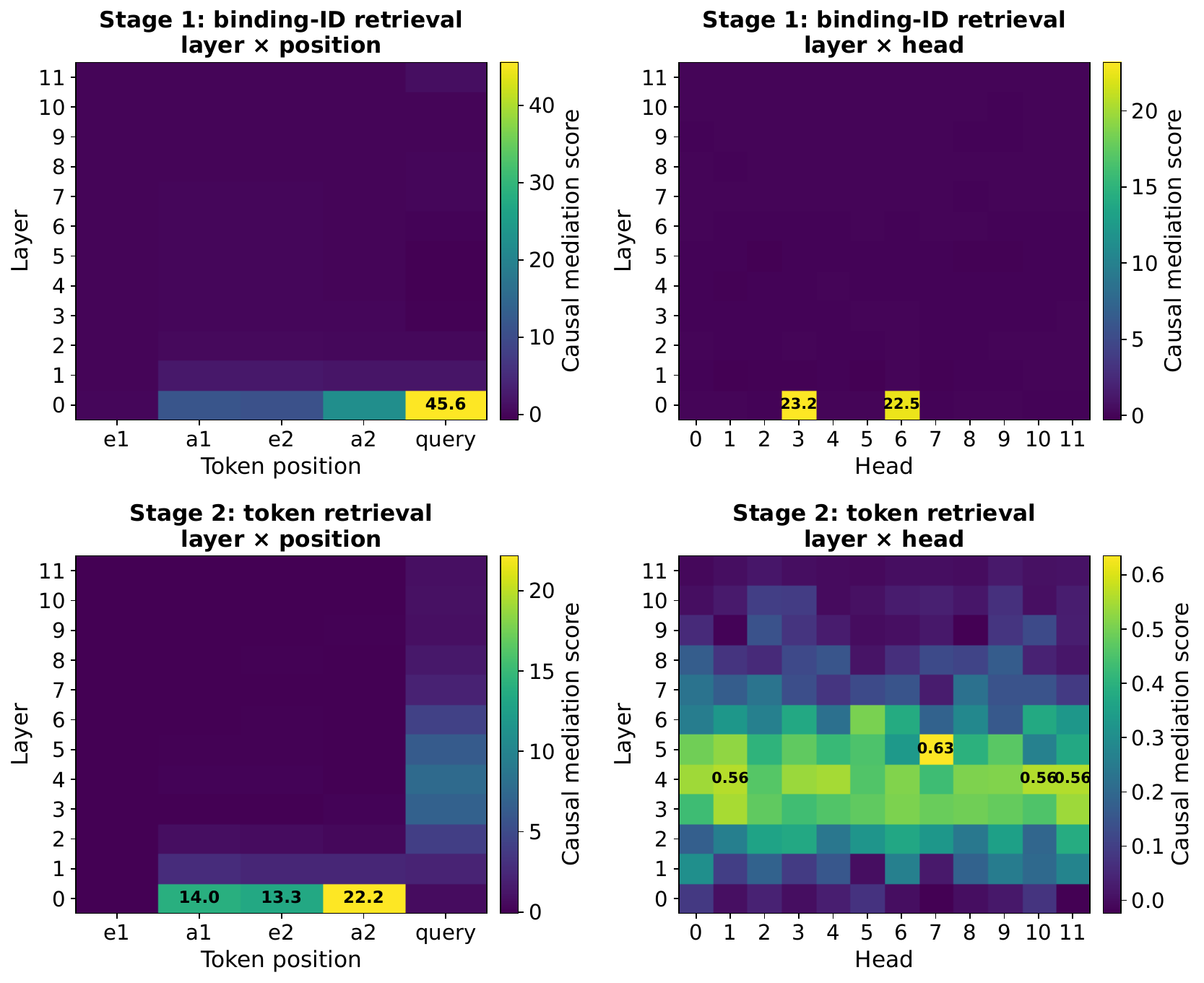}
    \caption{\textbf{Causal mediation analysis for binding.} \textit{Top:} Stage~1 (binding-ID retrieval). \textit{Bottom:} Stage~2 (token retrieval). \textit{Left:} layer $\times$ token position (patching attention and MLP output at one position). \textit{Right:} layer $\times$ head (patching one head's output at the query position). Layers are numbered from 0.}
    \label{fig:binding_cma}
\end{figure}
\subsection{Causal mediation analysis}
\label{app:binding_cma}

We use two patching experiments at the query position, each producing a score for every layer and token position and for every layer and attention head. In Stage~1 (binding-ID retrieval), the base prompt is $[e_1, a_1, e_2, a_2, e_1]$ and the patched prompt swaps which slot each pair occupies, $[e_2, a_2, e_1, a_1, e_1]$. Both have the same correct answer, $a_1$, because the entity-attribute binding is preserved; a head that reads the binding ID transfers the base prompt's slot and makes the patched prompt output $a_2$. In Stage~2 (token retrieval), the patched prompt swaps the attributes in place, $[e_1, a_2, e_2, a_1, e_1]$, so the correct answer changes; a head that reads the attribute token at the queried slot transfers the base prompt's attribute. A head's score is the change in the logit difference between the causal answer and the original answer when its output at the query position is patched from the base into the patched prompt.

For Stage~1 the effect is concentrated in layer 0 at the query position, and at the head level in two heads (layer 0, heads 3 and 6). For Stage~2 the effect is much more distributed: the head-level scores are spread over layers 2--6, strongest in layers 3--5 (peak at layer 5, head 7), and the layer-wise map additionally shows a large effect in layer 0 at the attribute positions. 

\subsection{Representational similarity analysis}
\label{app:binding_rsa}

To test what the identified heads represent, we computed per-head representational similarity at the query position. Each problem is rendered twice with the same tokens and the query fixed to the first entity: $[e_1, a_1, e_2, a_2, e_1]$ (the query's pair is in slot 0, ID0) and $[e_2, a_2, e_1, a_1, e_1]$ (slot 1, ID1). At the query position the token is identical across the pair while the binding ID differs, so a head that carries the ID gives a two-block similarity matrix (ID0 against ID0 and ID1 against ID1 similar, across-ID dissimilar), whereas a head that carries token identity gives a diagonal across the two blocks. We collected the head outputs for 20 problems (40 prompts), computed pairwise cosine similarities, and averaged the matrices over heads weighted by their CMA scores, keeping only heads whose score exceeds the mean by more than 0.5 standard deviations.

\begin{figure}[t]
    \centering
    \includegraphics[width=\textwidth]{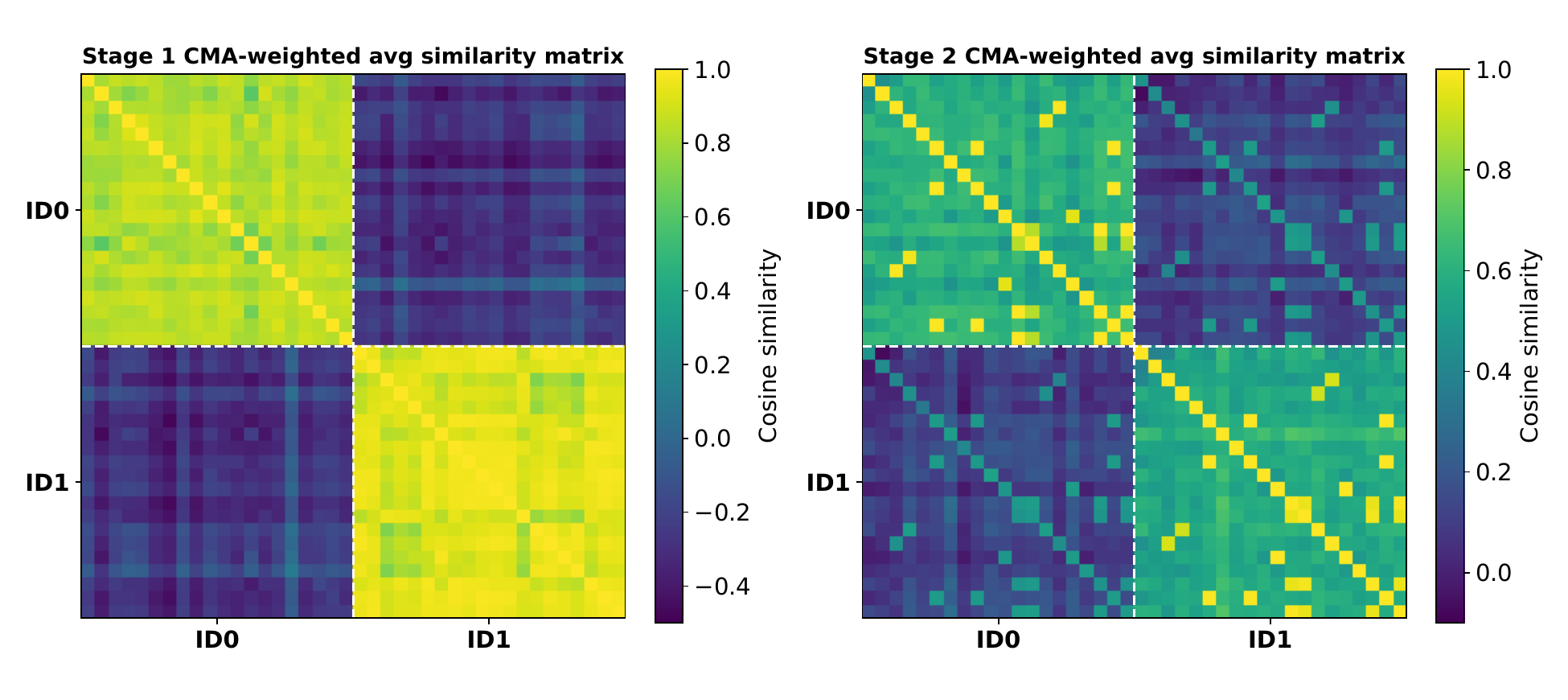}
    \caption{\textbf{CMA-weighted representational similarity for binding.} Average cosine-similarity matrices over heads with CMA $z>0.5$, ordered by binding ID (ID0 prompts first, ID1 second). \textit{Left:} Stage~1 (binding-ID retrieval). \textit{Right:} Stage~2 (token retrieval).}
    \label{fig:binding_rsa}
\end{figure}

The Stage~1 matrix had a strong two-block structure, which indicates that the Stage~1 heads encode the binding ID rather than token identity. The Stage~2 matrix keeps a block structure but also shows a diagonal linking the same token across the two IDs. 

\newpage

\end{document}